\pdfoutput=1

\documentclass[11pt]{article}

\usepackage[]{acl}

\usepackage{times}
\usepackage{latexsym}

\usepackage[T1]{fontenc}

\usepackage[utf8]{inputenc}

\usepackage{microtype}

\usepackage{inconsolata}
\usepackage{amsmath}
\usepackage{graphicx}
\usepackage{multicol}
\usepackage{subcaption}
\usepackage{booktabs}
\usepackage{multirow}
\usepackage{acl}
\usepackage[shortlabels]{enumitem}
\usepackage{times}
\usepackage{color,soul} 
\usepackage{booktabs}
\usepackage{enumitem}

\usepackage{latexsym}
\usepackage{hyperref}
\usepackage{multirow}
\usepackage[LAE,T1]{fontenc}
\usepackage{arabtex}
\usepackage{utf8} 
\usepackage[utf8]{inputenc} 
\usepackage[LAE,T1]{fontenc}
\usepackage[arabic,french,english]{babel}
\usepackage[raggedrightboxes]{ragged2e}
\usepackage{todonotes}
\usepackage{array}
\usepackage{lscape}
\usepackage{natbib}
\usepackage{longtable}
\usepackage{multirow}
\usepackage{float}
\usepackage{makecell}
\usepackage{xcolor, colortbl}
\usepackage{inconsolata}
\usepackage{arydshln}
\usepackage{array}
\usepackage{soul}
\usepackage{times}
\usepackage{latexsym}

\usepackage{multirow}

\usepackage{microtype}
\usepackage{hyperref}

\usepackage{microtype}
\usepackage{rotating}
\usepackage{amsmath}
\usepackage{amsfonts}
\usepackage{amssymb}

\usepackage{inconsolata}

\usepackage{graphicx}
\usepackage{subcaption}
\usepackage{multirow}
\usepackage{tabularx}
\usepackage{framed}

\usepackage{array} 

\title{What Do Speech Foundation Models Not Learn About Speech?}

\author{
  Abdul Waheed$^{1}$\,
  ~~~~~Hanin Atwany$^{2}$\,
  ~~~~~Bhiksha Raj$^{1,2}$\,
  ~~~~~Rita Singh$^{1}$\\
  $^{1}$Carnegie Mellon University~~~~~ 
  $^{2}$MBZUAI \\
  \texttt{\normalsize \{abdulw,bhiksha,rsingh\}@cs.cmu.edu}~~~~~~~~\texttt{\normalsize hanin.atwany@mbzuai.ac.ae}
}

\begin{document}
\maketitle
\section*{~~~~~~~~~~~~~~~~~~~~~~~~~~~~~Abstract}

Understanding how speech foundation models capture non-verbal cues is crucial for improving their interpretability and adaptability across diverse tasks. In our work, we analyze several prominent models—Whisper, Seamless, Wav2Vec, HuBERT, and Qwen2-Audio—focusing on their learned representations in both paralinguistic and non-paralinguistic tasks from the Dynamic-SUPERB benchmark. Our study addresses three key questions: (i) what non-verbal cues (e.g., speaker intent, emotion, environmental context) are captured? (ii) how are these cues represented across different layers of the models? and (iii) to what extent can these representations be effectively adapted to downstream tasks?
To answer these questions, we first evaluate the models in a zero-shot setting, followed by fine-tuning on layer-wise features extracted from these models. Our results provide insights into the models' capacity for generalization, the characteristics of their layer-wise representations, and the degree of transformation required for downstream task adaptation. Our findings suggest that some of these models perform well on various tasks in zero-shot settings, despite not being explicitly trained for those tasks. We also observe that zero-shot performance correlates with better-learned representations. The analysis of layer-wise features demonstrates that some models exhibit a convex relationship between the separability of the learned representations and model depth, with different layers capturing task-specific features.

\section{Introduction}~\label{sec:introduction}

The rise of large language models (LLMs) using unsupervised pre-training has revolutionized natural language processing (NLP)~\cite{bubeck2023sparksartificialgeneralintelligence,minaee2024largelanguagemodelssurvey, liu2024understanding,kaddour2023challengesapplicationslargelanguage}. This paradigm shift has spurred an increased focus on understanding the inner workings and capabilities of these models~\cite{ju2024large,jin2024exploring,van2023exploration,prakash2023layered}. While initially developed for text-based tasks, transformer-based architectures~\cite{vaswani2023attentionneed} have since been adapted for speech processing tasks, such as automatic speech recognition (ASR), text-to-speech (TTS), and speech synthesis~\cite{wang2019overview,baevski2020wav2vec20frameworkselfsupervised}.

Despite considerable research exploring the knowledge captured within text-based models~\cite{ferrando2024primerinnerworkingstransformerbased, mosbach2024insightsactionsimpactinterpretability, hassanin2024comprehensive, rogers2020primerbertologyknowbert,luo2024understandingutilizationsurveyexplainability,räuker2023transparentaisurveyinterpreting,sindhu2024evolution,raiaan2024review,jawahar2019does}, our understanding of speech foundation models~\footnote{A foundation model refers to a large-scale, pre-trained model that serves as a general-purpose model for a wide range of tasks. } remains limited. Although these models have demonstrated remarkable performance across a variety of tasks, including ASR, speaker identification, and emotion detection, there has been little investigation into what these models actually learn and how their internal representations correspond to various speech-related tasks~\cite{tierney2020speech,10446502, cui2024recentadvancesspeechlanguage}.

The black-box nature of these models raises critial questions about whether they capture the rich, nuanced information encoded in human speech—such as paralinguistic features like emotion, stress, or environmental context—or if their impressive results are driven mainly by surface-level patterns. Addressing this gap is crucial not only for improving model interpretability but also for guiding the development of more robust and generalizable speech models. Therefore, it is essential to systematically examine speech foundation models to uncover the representations they learn and better understand how these representations align with a diverse set of speech-related tasks.

In this work, we aim to address this research gap by conducting a comprehensive analysis of several speech foundation models, including Whisper~\cite{radford2023robust}, Seamless~\cite{barrault2023seamlessm4t}, Wav2Vec~\cite{Schneider2019wav2vecUP}, HuBERT~\cite{9585401}, and Qwen2-Audio~\cite{chu2023qwen}. Our investigation focuses on comparing the discriminative characteristics of their learned representations with handcrafted audio features extracted using Librosa~\cite{mcfee2015librosa}\footnote{\url{https://librosa.org/doc/latest/feature.html}}. Additionally, we explore how well these models generalize to new tasks through zero-shot evaluations, providing insights into their performance on unseen tasks.

More specifically, we select ten tasks from the Dynamic-SUPERB~\cite{10448257} benchmark, encompassing both paralinguistic and non-paralinguistic tasks. For each model, we extract features from each layer and train K-Nearest Neighbors (KNN)~\cite{peterson2009k} and Neural Networks (NN)~\cite{yi2016study} classifiers to evaluate and compare their performance. We also conduct zero-shot evaluations with various prompts to assess the models' generalization capabilities.

Through these experiments, we aim to uncover what these models learn beyond verbal content, focusing on their ability to interpret non-verbal cues. By emphasizing non-content-based tasks, we evaluate how well the models capture paralinguistic features like speaker intent, mood, and context. This helps us assess whether the models grasp the subtleties of human speech, where non-verbal signals are as important as spoken words. Including these tasks provides valuable insights into the models' ability to generalize beyond basic transcription and synthesis, allowing us to measure their understanding of the more nuanced aspects of speech.

\noindent We summarize our contributions as follows:
\begin{itemize}
    \item We conduct a zero-shot evaluation of various speech foundation models across a diverse range of paralinguistic and non-paralinguistic tasks.
    \item We extract layer-wise features from these models and train classifiers to assess the discriminative capabilities of the learned representations.
    \item We provide a thorough analysis of our findings, showing that for many tasks and models, there exists a convex relationship between model performance and specific regions within the model's layers.
\end{itemize}

\noindent\textbf{Outline.} The remainder of the paper is organized as follows: In Section~\ref{sec:related-work}, we present a review of existing literature. Section~\ref{sec:prelimnaries} discusses preliminaries. In Section~\ref{sec:experiments}, we detail the experimental setup, while Section~\ref{sec:results} covers the results obtained and offers a discussion of the findings. and we conclude in Section~\ref{sec:conclusion}. The limitations of our work are outlined in Section~\ref{sec:limitations}.

\section{Related Work}~\label{sec:related-work} 

Recent advancement in building speech foundation models~\cite{zhang4725202whisper, barrault2023seamless, pratap2023scalingspeechtechnology1000, zhang2023googleusmscalingautomatic,li2023efficient, baevski2020wav2vec20frameworkselfsupervised} has resulted in massive improvement on downstream tasks such as speech-recognition~\cite{10445874} and speech-to-text translation~\cite{zhang4725202whisper}. In addition, these models also show the ability to generalize to novel and unseen tasks~\cite{yang2021superb}. However, their understanding of non-verbal cues in speech remains unexplored \cite{martin2023speaking}, and an analysis of the representations they learn has yet to be investigated \cite{sanabria2023analyzing}.

A number of studies have assessed the transferability of speech model representations for cross-task downstream speech tasks \cite{chemudupati2023transferability,guimaraes2023exploration,chen2023estimate,barrault2023seamlessm4t}. These works aim to interpret the knowledge encoded in extracted features to address the black-box nature of these models \cite{liu2021probing,belinkov2020interpretability}. For this purpose, different methods and frameworks have been developed, including probing classifiers, studies on language modeling behavior~\cite{belinkov2022probing}, inference tasks~\cite{liu2024large}, psycholinguistic approaches~\cite{trott2024can}, layerwise analyses~\cite{ju2024large}, and other relevant methodologies \cite{wang2018can, hewitt2019structural}.

Through probing tasks, researchers found that models tend to encode richer contextual information in the upper layers, expanding from entity-level knowledge in the lower layers~\cite{ju2024large, chowdhury2023endtoendspeechmodelslearn, Jawahar2019WhatDB, RameshKashyap2021AnalyzingTD}. Previous work, such as~\cite{jawahar2019does}, has explored what BERT learns about language structure, revealing that lower layers capture phrase-level information, while intermediate layers encode syntactic and semantic features. However, most studies have focused on language models, leaving speech foundation models largely unexplored.

Speech models like Whisper~\cite{zhang4725202whisper}, Seamless~\cite{barrault2023seamlessm4t}, Wav2Vec2~\cite{tom2022wav2vec}, HuBERT~\cite{hsu2021hubertselfsupervisedspeechrepresentation}, and Qwen2-Audio~\cite{chu2023qwen} leverage different architectures and training methods to improve speech understanding and ASR. While Whisper has shown potential beyond ASR with its vast multilingual dataset \cite{goron2024improving}, Wav2Vec2 and HuBERT focus on self-supervised learning. Despite these advances, little is known about the specific knowledge encoded in each layer of these models. Our work fills this gap by evaluating layer-wise features across tasks using NN and KNN classifiers to identify the most effective layers for task-specific performance.

\section{Preliminaries}~\label{sec:prelimnaries}

In this section, we present the two primary approaches for evaluating speech foundation models: zero-shot learning with text prompts and feature-based fine-tuning using k-Nearest Neighbors (KNN) and Neural Networks (NNs).

\subsection{Zero-Shot Evaluation}~\label{subsec:zero-shot}

In the zero-shot setting, we evaluate the models' ability to generalize to new tasks by providing speech input and a text prompt describing the task, without any task-specific fine-tuning. Let $\mathcal{M}$ be a speech foundation model trained on a large dataset $\mathcal{D} = \{(\mathbf{x}_i, \mathbf{y}_i)\}_{i=1}^N$, where $(\mathbf{x}_i, \mathbf{y}_i)$ represent speech inputs and corresponding labels. For zero-shot evaluation, we input test speech $\mathbf{x}_{test}$ along with a task-specific prompt $\mathbf{p}_{task}$ (e.g., transcription or speaker classification). The model outputs the log probabilities for each class.

Let $\mathcal{L}_{c}$ represent the set of possible classes for the given task. The model $\mathcal{M}$ computes the log probability $\log P(c|\mathbf{x}_{test}, \mathbf{p}_{task})$ for each class $c \in \mathcal{L}_{c}$. The predicted class $\mathbf{\hat{y}}_{test}$ is selected as the one with the highest log probability:

\[
\mathbf{\hat{y}}_{test} = \arg \max_{c \in \mathcal{L}_{c}} \log P(c|\mathbf{x}_{test}, \mathbf{p}_{task})
\]

This approach assesses how well the pre-trained model generalizes to unseen tasks using its learned representations. The task-specific knowledge is embedded in the text prompt, and the model uses its internal representations to infer the most likely class based on the speech input and task description.

\subsection{Fine-Tuning on Features}~\label{subsec:fine-tuning}

In addition to zero-shot evaluation, we analyze the models by extracting features from each layer and training classifiers on top of these features. Let $\mathcal{F}_l(\mathbf{x}_i)$ denote the feature representation extracted from the $l$-th layer of the model $\mathcal{M}$ for input $\mathbf{x}_i$. We extract layer-wise features from all layers $l \in \{1, \dots, L\}$, where $L$ is the total number of layers in the model.


For KNN, the classifier finds the $k$ nearest neighbors of the feature representation $\mathbf{z}_i^l = \mathcal{F}_l(\mathbf{x}_i)$ in the feature space and assigns a label $\mathbf{\hat{y}}_i^{KNN}$ based on majority voting among the neighbors. The prediction for the KNN model is given by:

\[
\mathbf{\hat{y}}_i^{KNN} = \arg \max_c \sum_{j \in \mathcal{N}_k(\mathbf{z}_i^l)} \mathbb{1}[\mathbf{y}_j = c]
\]

where $\mathcal{N}_k(\mathbf{z}_i^l)$ is the set of $k$ nearest neighbors of $\mathbf{z}_i^l$ in the feature space, and $\mathbb{1}[\mathbf{y}_j = c]$ is an indicator function that checks whether the label of neighbor $j$ matches class $c$.

For the NN classifier, we define a learnable function $\mathcal{G}_\theta(\mathbf{z}_i^l)$ parameterized by $\theta$, which maps the extracted feature vector $\mathbf{z}_i^l$ to class predictions. The NN model is trained by minimizing the cross-entropy loss:

\[
\mathcal{L}_{DNN} = - \frac{1}{N} \sum_{i=1}^{N} \sum_{c=1}^{C} \mathbf{y}_i^c \log \mathcal{G}_\theta(\mathbf{z}_i^l)
\]

where $C$ is the number of classes, and $\mathbf{y}_i^c$ is the one-hot encoded ground truth label for class $c$. The parameters $\theta$ are optimized to minimize the loss, allowing the NN to fine-tune the mapping from the extracted features to the class labels.


Training both KNN and NN classifiers on the extracted features is crucial for understanding the quality of the learned representations at different model layers. KNN, as a non-parametric model, reflects the discriminative power of the raw features without learning any weights, providing insight into how well the features capture similarities in an unsupervised manner. NNs, with their learnable parameters, allow for more complex, task-specific transformations, improving performance when features are not linearly separable. 

Thus, evaluating both classifiers offers a comprehensive view of feature quality—KNN reveals raw feature utility, while NN measures the potential for further task-specific adaptation.


\section{Experiments}~\label{sec:experiments}

We outline our experimental framework designed to evaluate the performance of various foundational speech models across ten distinct audio classification tasks. Our analysis is structured to assess the impact of different models, focusing on their learned representations and performance at each layer.

\subsection{Dataset}
We carefully select a subset of tasks from the Dynamic-SUPERB benchmark~\cite{10448257}, which spans a diverse range of tasks related to speech understanding. While these foundation models have demonstrated strong performance on content-based tasks such as speech-to-text and text-to-speech~\cite{barrault2023seamlessm4t, radford2023robust}, we limit our evaluation to tasks that are not content-based, following the categorization outlined in~\cite{10448257, yang2021superb}. This decision is essential because focusing on non-content-based tasks, such as paralinguistic and non-paralinguistic tasks, helps us better understand what these speech foundation models are learning beyond verbal content—specifically, their ability to interpret non-verbal cues.

Non-verbal cues, such as emotion, speaker identity, and environmental sounds, are crucial in real-world communication systems \cite{dzardanova2024exploring}. By emphasizing these tasks, we can evaluate how effectively the models capture and utilize paralinguistic features that convey speaker intent, mood, and context. This understanding is key to assessing whether the models truly comprehend the subtleties of human communication, where non-verbal signals often play an equally important role as linguistic information.

We choose ten tasks, broadly categorizing them into paralinguistic and non-paralinguistic groups. Each task in Dynamic-SUPERB comes with clearly defined instructions, facilitating the effective evaluation of model performance in previously unseen contexts. Further details on the tasks used in our evaluation are provided in Table~\ref{tab:tasks}.


\begin{table}[h!]
\centering
\begin{tabular}{llc}
\toprule
\textbf{Task} & \textbf{Type} & \textbf{nClass} \\ \midrule
Accent Class. & Paraling & 9\\
Dialogue Act & Semantic & 4\\
Emotion Recog. & Paraling & 7\\
Env. Sound Class. & Audio  & 10\\
HowFarAreYou & Paraling & 3\\
Intent Class. & Semantic & 6\\
Multi-Speaker Det. & Speaker & 2\\
Sarcasm Det. & Paraling & 2\\
Spoof Det. & Paraling & 2\\
Stress Det. & Paraling & 6 \\ \bottomrule
\end{tabular}
\caption{Table of tasks and their respective categories in the Dynamic-SUPERB benchmark, along with the number of target classes (nClass) for each task.}
\label{tab:tasks}
\end{table}

\subsection{Models}
For a more inclusive study, we select three types of models: encoder-only, decoder-only, and encoder-decoder. Each model varies in training data, objectives, and architecture, providing unique insights into how these foundation models process and understand speech.

Encoder-only models (e.g., HuBERT~\citet{hsu2021hubertselfsupervisedspeechrepresentation}, Wav2Vec~\citet{schneider2019wav2vec}) specialize in extracting acoustic and paralinguistic features, making them ideal for tasks like speaker identification, emotion recognition, and environmental sound classification.

Decoder-only models (e.g., Qwen2-Audio \cite{chu2024qwen2}), typically employed for autoregressive tasks such as speech-to-text, enable us to evaluate their capability to manage paralinguistic elements beyond mere sequence generation.

Encoder-decoder models (e.g., Whisper~\citet{radford2023robust}, SeamlessM4T~\cite{barrault2023seamlessm4t}) combine both feature extraction and sequence generation, making them versatile for a variety of tasks that involve both understanding and generating speech.

By selecting models with distinct architectures, objectives, and training data, we aim explore their capabilities across a broad range of tasks, assessing performance on both paralinguistic and non-paralinguistic challenges. Further details about these models are provided in Table~\ref{tab:models}.

\begin{table}[h!]
\begin{tabular}{lll}
\toprule
Type            & Models            & Evaluation \\ \midrule
Enc         & HuBERT, Wav2Vec  & FT         \\
Enc-Dec & Whisper, Seamless & ZS, FT     \\
Dec         & Qwen2-Audio        & ZS, FT    \\ \bottomrule
\end{tabular}
\caption{Description of the models we used in our evaluation. Abbreviation: Enc - Encoder, Dec - Decoder, ZS - Zero-shot, FT -  Finetunning}
\label{tab:models}
\end{table}

\subsection{Training and Evaluation}~\label{subsec:training}

\noindent\textbf{Zero-Shot.}~\label{text:zero-shot}
In the zero-shot setting, we evaluate the ability of models, specifically decoder-only and encoder-decoder models, to generalize to new tasks without any task-specific fine-tuning. These models are capable of generating text from speech input, making them suitable for zero-shot evaluation across various tasks. 

For this evaluation, we provide the model with speech input and a text prompt that describes the task (e.g., emotion recognition, speaker classification). Without any prior exposure to the specific task during fine-tuning, the model is expected to infer the correct output based on its pre-trained knowledge. The text prompt helps guide the model's understanding of the task, while the model leverages its internal representations to process the speech input and produce the corresponding output.

The zero-shot approach allows us to assess how well these models can handle unseen tasks by using the knowledge embedded in their pre-trained representations. Our setup is particularly important for understanding how flexible and adaptable the models are when faced with new, previously unseen scenarios.
In our experiments which is also supported by the findings in~\cite{10448257}, we find that zero-shot models are susceptible to prompt and as result of this we compare three different prompts for zero-shot evaluation. We provide more details about our prompts in Table~\ref{tab:prompts}.


\begin{table}[]
\setlength{\tabcolsep}{4pt} 
\begin{tabular}{lp{6cm}}
\toprule
\textbf{Name}   & \textbf{Prompt}                                                                                                     \\ \midrule
MCQ & \{instruction\}. Choose the correct answer: \{options\}. Your answer: \\ \hdashline
Quiz   & \{instruction\}. Which of the following options is the correct answer? \{options\}. \\ \hdashline
Blank  & \{instruction\}. The correct label is \_\_\_\_. Choose from: \{options\}. Answer: \\ \bottomrule
\end{tabular}
\caption{Prompts used for zero-shot evaluation.}
\label{tab:prompts}
\end{table}

\noindent\textbf{Supervised Finetuning.}~\label{supervised-finetuning}
We train  K-Nearest Neighbors (KNN) and Neural Networks (NN) classifiers on audio feature representations extracted from speech foundation models. Both encoder and decode features from each model were used, evaluating classification performance across individual layers as well as mean-pooled features. We standardize the representation using \texttt{StandardScaler} to ensure numerical stability during training. 
For both KNN and NN, We used 5-fold stratified cross-validation for all tasks. 
The primary evaluation metric utilized in our experiments was the F1 score, which measures the harmonic mean of precision and recall across all classes for each task. We computed results using k-fold cross-validation, averaging the Macro-F1 scores across folds and reporting standard deviations to quantify the variability and robustness of the results.

\subsection{Setup}

We utilize Huggingface Transformers~\cite{wolf2020huggingfacestransformersstateoftheartnatural}~\footnote{https://huggingface.co/docs/transformers/en/index} to load the pre-trained speech foundation models. Depending on model size, we run our experiments on either 1xA100 GPU or 4xA100 GPUs (for larger models). For each model, we extract layerwise features, which are then used to train Neural Networks (NN) and run K-Nearest Neighbors (KNN) classifiers, as described in~\ref{supervised-finetuning} and in~\ref{sec:prelimnaries}.

We provide specific hyperparameters used in our experiments in Appendix~\ref{appendix-subsec:hyperparameters}.

\section{Results and Discussion}~\label{sec:results}
\begin{table*}[!hbt]
\centering
\resizebox{\textwidth}{!}{%
\begin{tabular}{lcccccccc}
\toprule
\textbf{Task} & \textbf{W-L-v3} & \textbf{D-L-v3} & \textbf{W-M} & \textbf{W-M.en} & \textbf{SM4T-M} & \textbf{SM4T-v2-L} & \textbf{Qwen2} & \textbf{Qwen2-I} \\
\midrule
AccentClassification & 5.30 & 6.90 & 6.27 & 8.02 & 10.57 & 1.28 & 3.95 & 4.04 \\
DialogueActClassification & 18.27 & 19.74 & 16.52 & 17.97 & 23.27 & 9.16 & 26.37 & 23.98 \\
EmotionRecognition & 4.64 & 1.70 & 8.15 & 5.71 & 7.33 & 2.59 & 19.70 & 11.53 \\
EnvironmentalSoundClassification & 4.55 & 6.74 & 8.27 & 15.06 & 9.93 & 1.65 & 28.46 & 6.86 \\
HowFarAreYou & 29.41 & 33.51 & 31.12 & 26.49 & 28.41 & 16.66 & 24.94 & 26.22 \\
IntentClassification & 19.94 & 16.42 & 25.95 & 24.11 & 35.61 & 3.95 & 30.14 & 23.24 \\
MultiSpeakerDetection & 49.40 & 58.63 & 35.87 & 52.71 & 45.91 & 33.48 & 50.22 & 41.42 \\
SarcasmDetection & 36.33 & 47.09 & 44.99 & 49.12 & 42.86 & 35.06 & 35.06 & 41.65 \\
SpoofDetection & 50.17 & 45.79 & 47.76 & 54.82 & 48.62 & 34.46 & 47.60 & 8.41 \\
StressDetection & 14.48 & 13.40 & 13.32 & 10.31 & 13.84 & 2.57 & 9.42 & 6.25 \\
\midrule
\textbf{Average} & 23.25 & 24.99 & 23.82 & 26.43 & 26.64 & 14.09 & 27.59 & 19.36 \\
\bottomrule
\end{tabular}%
}
\caption{F1 score for zero-shot evaluation, averaged across three prompts. Abbreviations: W - Whisper, L - Large, D - DistilWhisper, M - Medium, SM4T - SeamlessM4T, I - Instruct.}
\label{tab:zero-shot-results}
\end{table*}

We conduct two primary types of experiments. First, we assess various Speech Foundation Models (SFMs) in a zero-shot setting. Next, we extract layer-wise features from all models and train KNN and NN classifiers using these features. This process is applied across all ten tasks in our study. The following sections detail the results of these experiments.

\subsection{Zero-Shot Results}

In our experiments, we evaluate various SFMs in a zero-shot setting. For this, we test each model's ability to generalize to different tasks without fine-tuning, as described in~\ref{subsec:zero-shot}.

Our findings reveal that while some models perform well on several downstream tasks, their zero-shot performance on other tasks can be worse than the random classifier baseline. For instance, Whisper-large-v3 performs better than random classification in many tasks where it was not explicitly trained, demonstrating its ability to generalize to new tasks in zero-shot setting. However, for more complex (more classes) tasks like \textit{Stress Detection} and \textit{Emotion Recognition}, even models such as Whisper fall short, highlighting challenges in generalizing across all task types without fine-tuning. We show zero-shot results in Table~\ref{tab:zero-shot-results}.

We notice that soome models are sensitive to prompt type, with performance varying across tasks. For instance, Whisper-large-v2 performs best with MCQ prompts for Dialogue Act and Spoof Detection, while fill-in-the-blank works better for Environmental Sound Classification. Distil-large-v3 shows consistent performance, indicating lower prompt sensitivity. In contrast, Whisper-medium.en is highly sensitive, especially in tasks like Intent Classification and Spoof Detection. Detailed results for each are in Appendix Table~\ref{tab:speech-model-performance}.


\subsection{Classification Results on Layer-wise Features}
\begin{figure*}[htbp]
    \centering
    \begin{subfigure}[b]{0.3\textwidth}
        \centering
        \includegraphics[width=\textwidth]{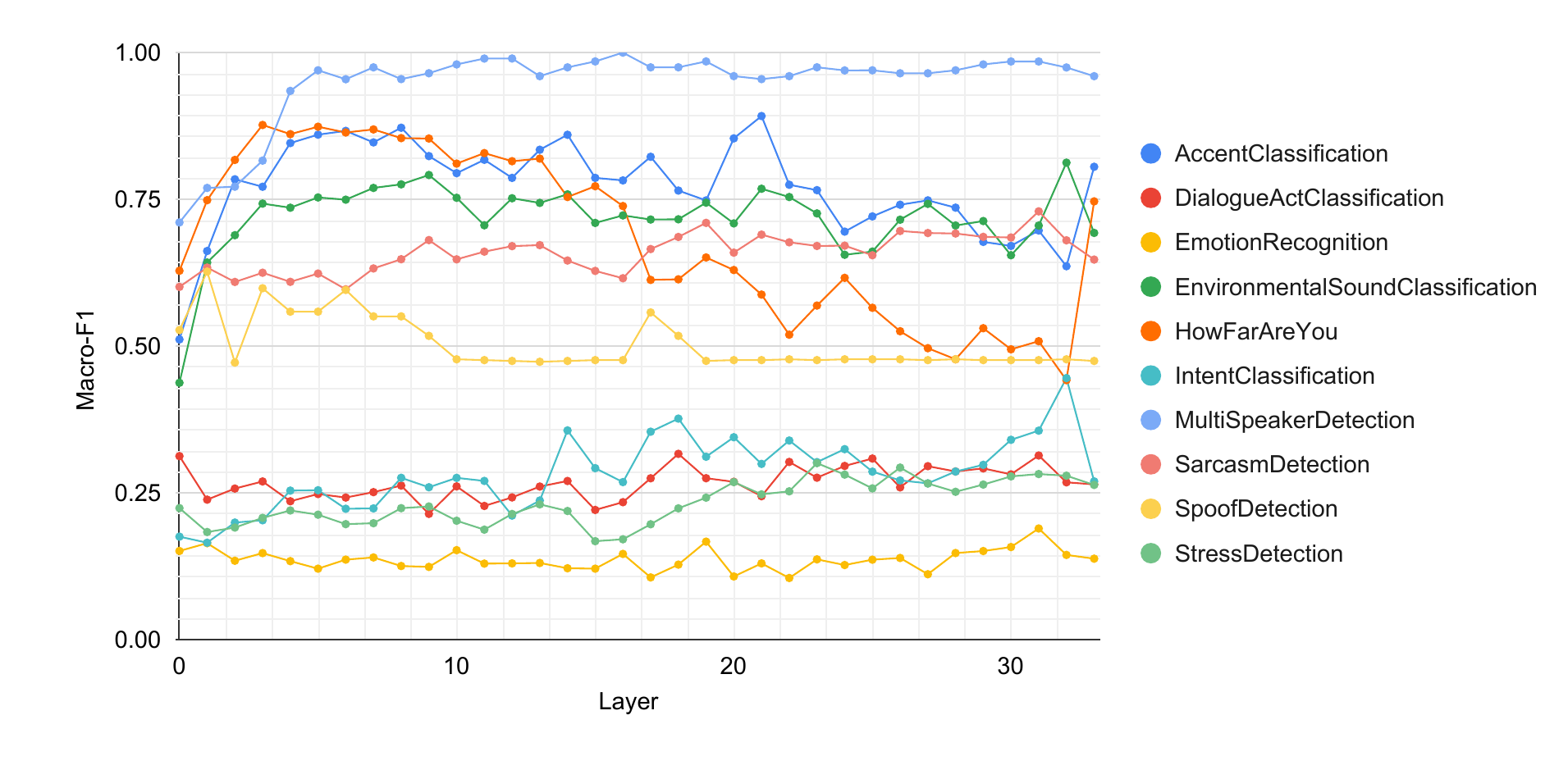}  
        \caption{Whisper-large-v3}
        \label{fig:first}
    \end{subfigure}
    \hspace{0.03\textwidth}  
    \begin{subfigure}[b]{0.3\textwidth}
        \centering
        \includegraphics[width=\textwidth]{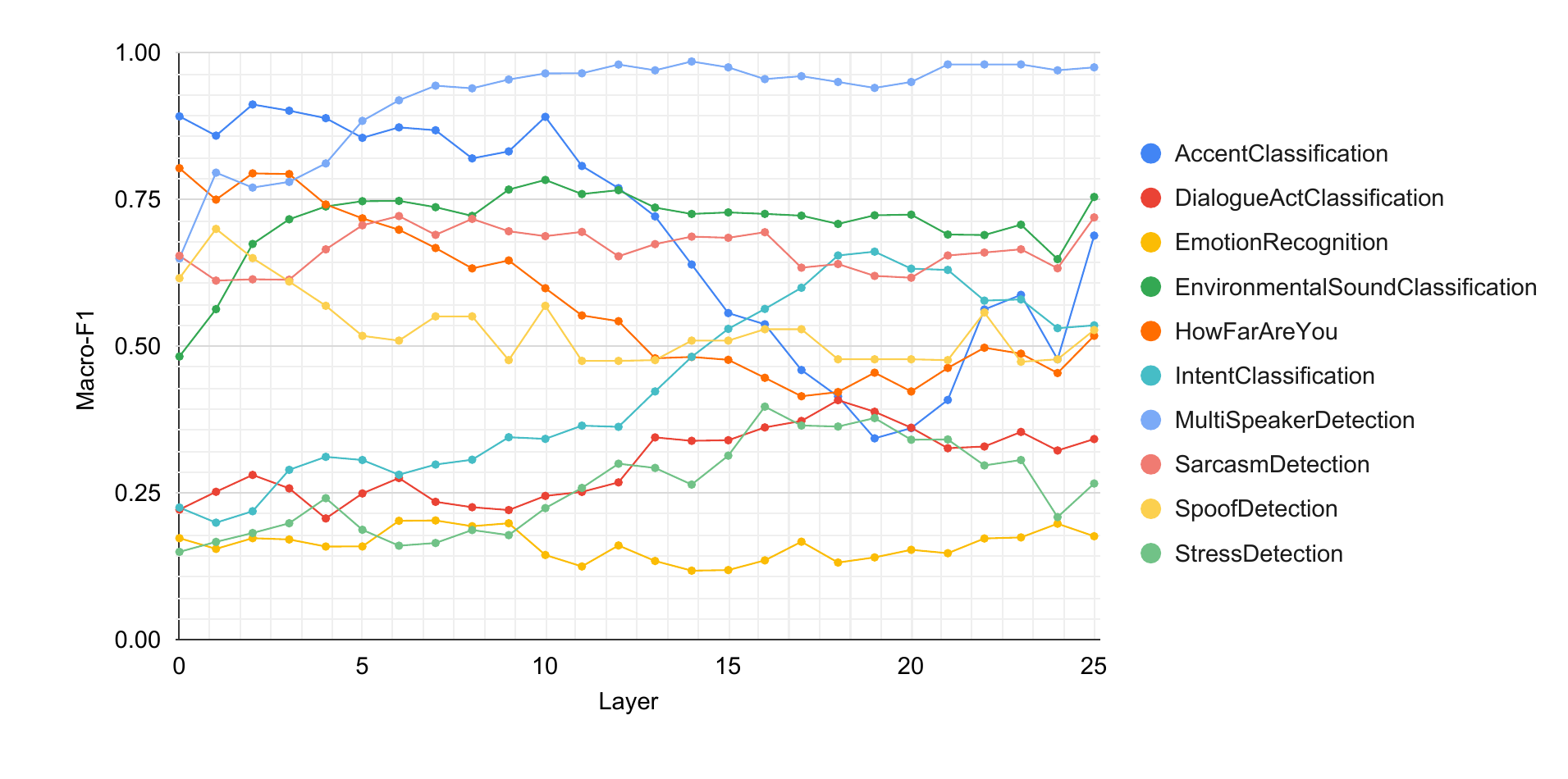}  
        \caption{HuBERT}
        \label{fig:second}
    \end{subfigure}
    \hspace{0.03\textwidth}  
    \begin{subfigure}[b]{0.3\textwidth}
        \centering
        \includegraphics[width=\textwidth]{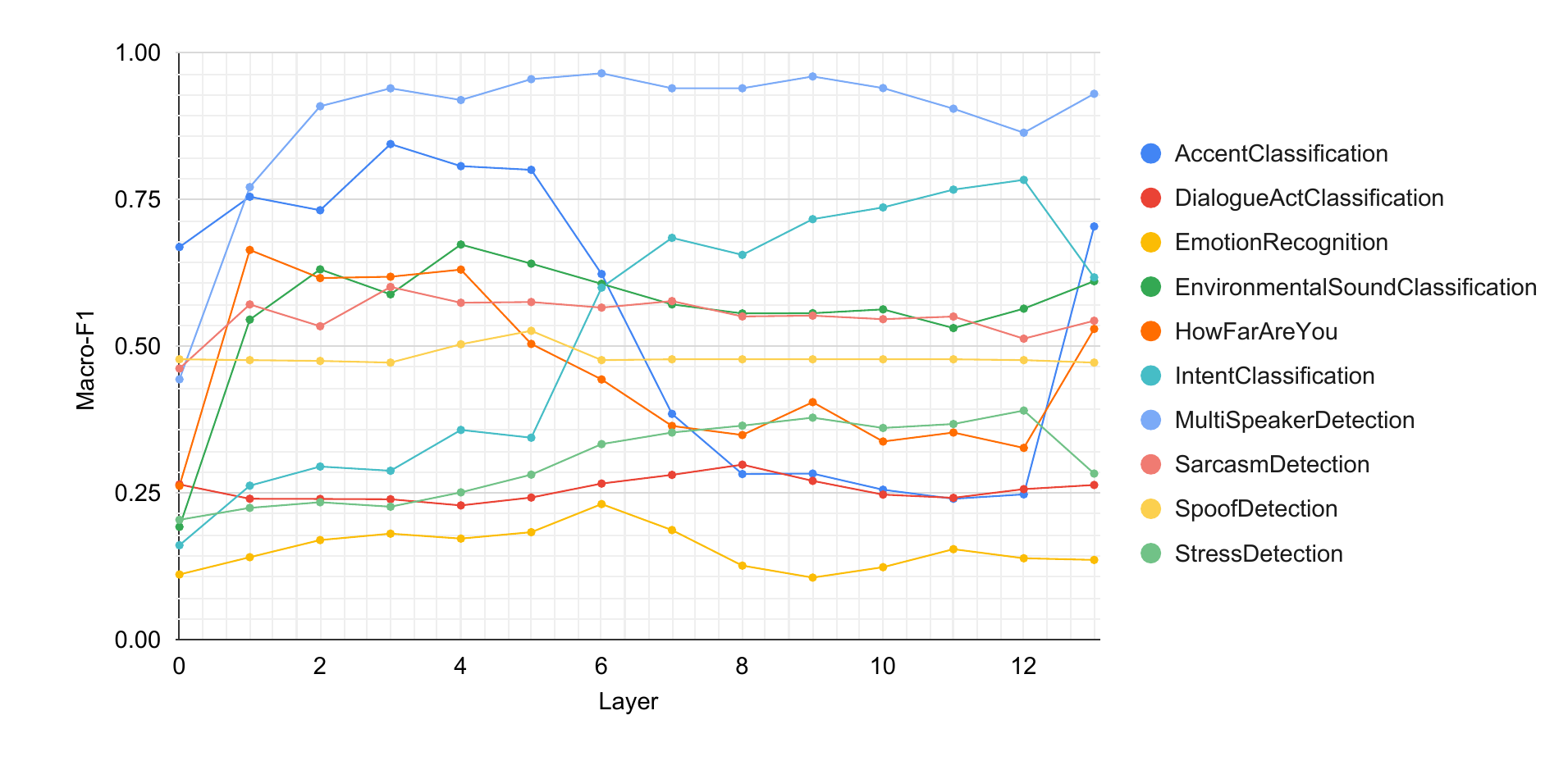}  
        \caption{Seamless-medium}
        \label{fig:third}
    \end{subfigure}
    
    \vspace{0.5cm}  
    \begin{subfigure}[b]{0.3\textwidth}
        \centering
        \includegraphics[width=\textwidth]{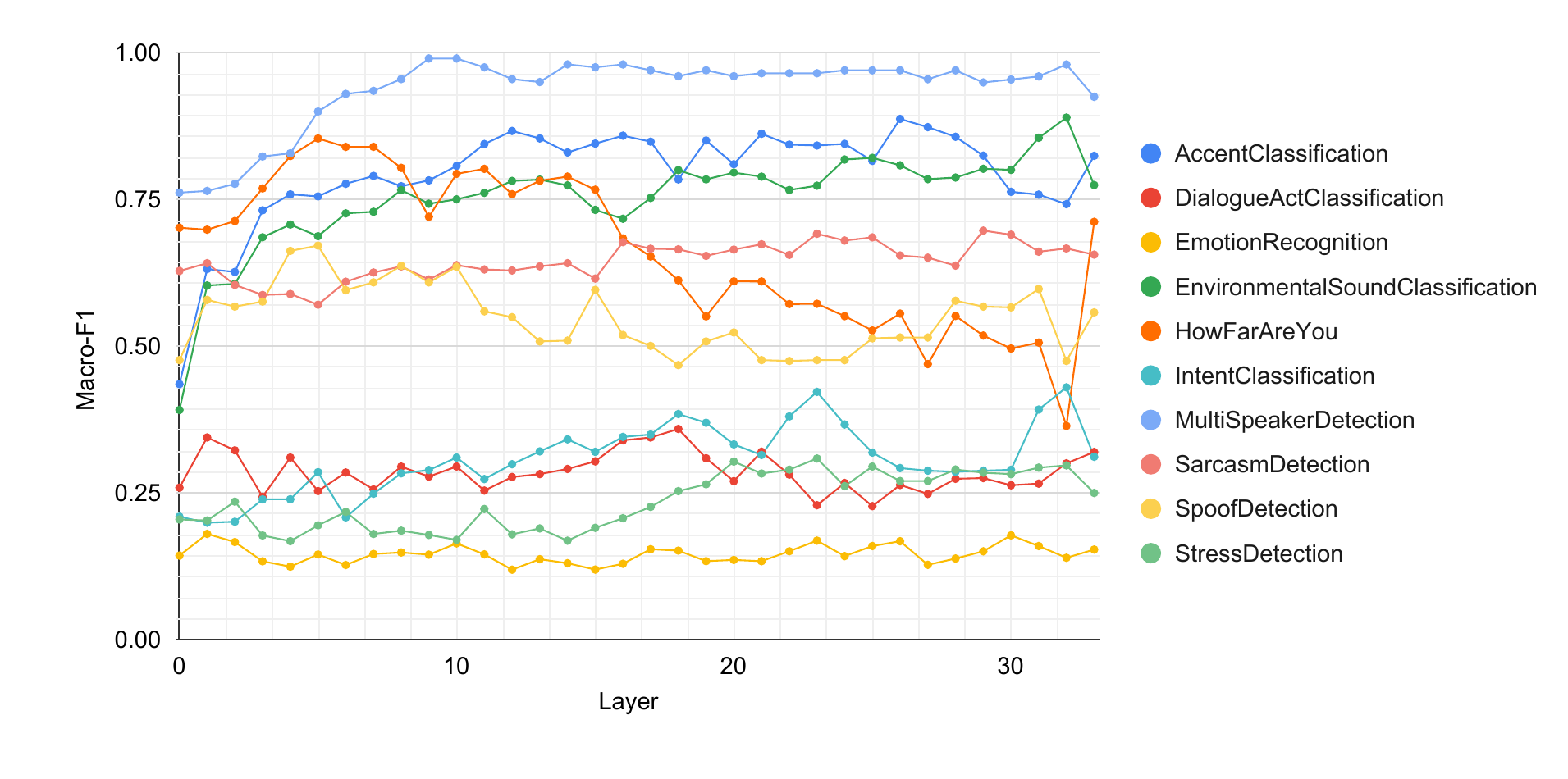}  
        \caption{Whisper-large-v2}
        \label{fig:fourth}
    \end{subfigure}
    \hspace{0.03\textwidth}  
    \begin{subfigure}[b]{0.3\textwidth}
        \centering
        \includegraphics[width=\textwidth]{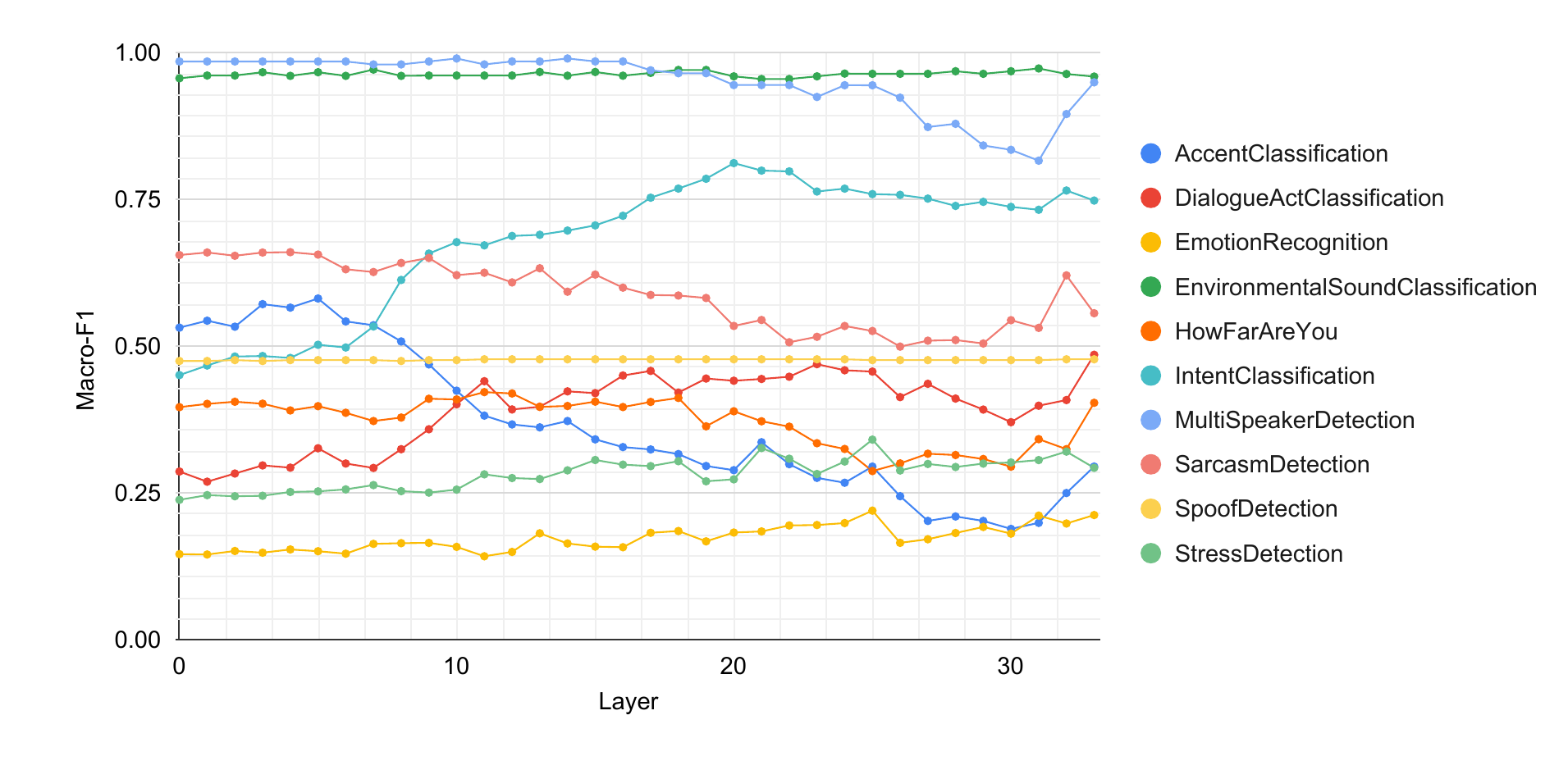}  
        \caption{Qwen2-Audio-Instruct}
        \label{fig:fifth}
    \end{subfigure}
    \hspace{0.03\textwidth}  
    \begin{subfigure}[b]{0.3\textwidth}
        \centering
        \includegraphics[width=\textwidth]{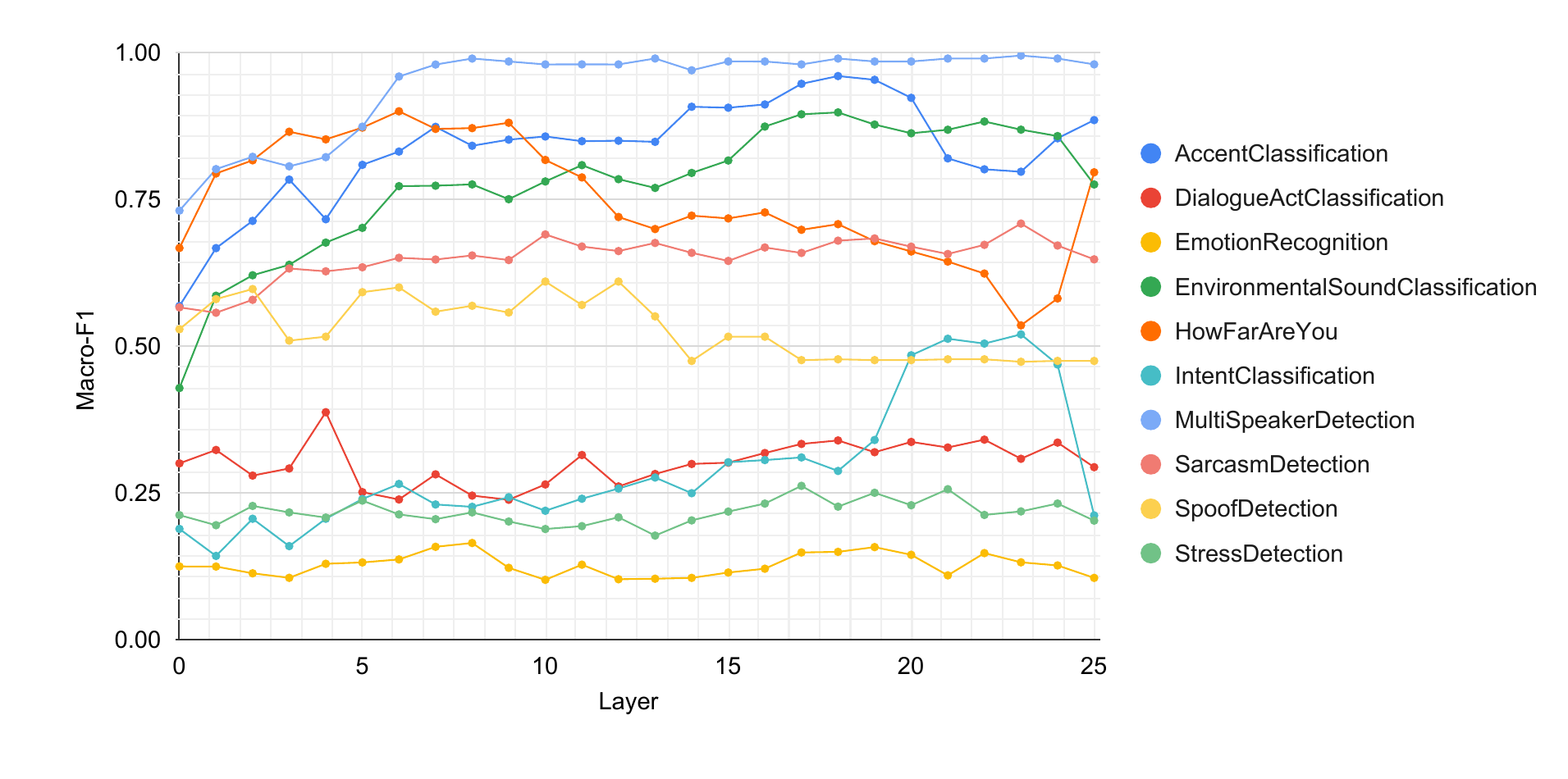}  
        \caption{Whisper-medium-en}
        \label{fig:sixth}
    \end{subfigure}

    \vspace{0.5cm}  
    \begin{subfigure}[b]{0.3\textwidth}
        \centering
        \includegraphics[width=\textwidth]{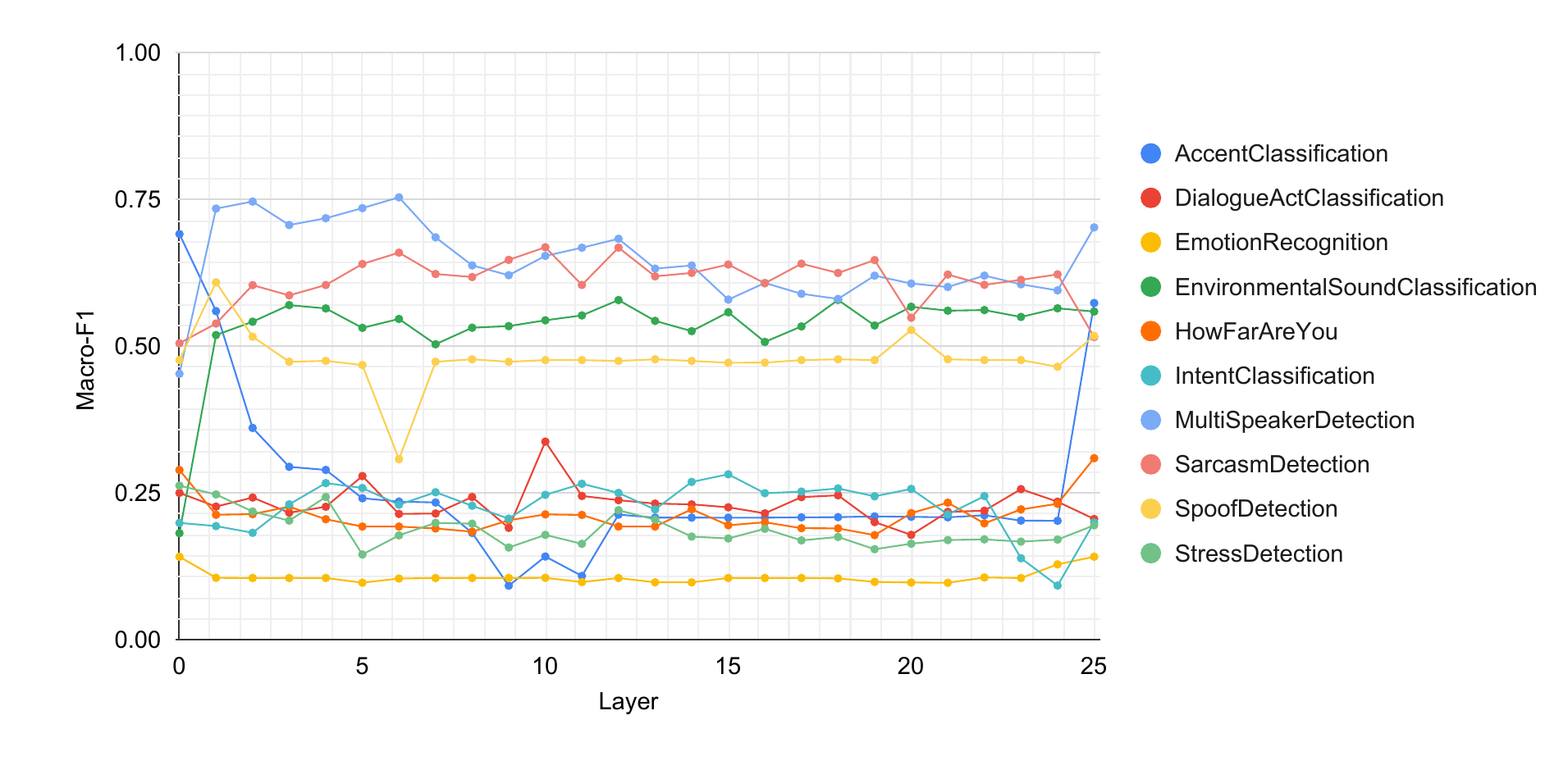}  
        \caption{Seamless-v2-large}
        \label{fig:fourth}
    \end{subfigure}
    \hspace{0.03\textwidth}  
    \begin{subfigure}[b]{0.3\textwidth}
        \centering
        \includegraphics[width=\textwidth]{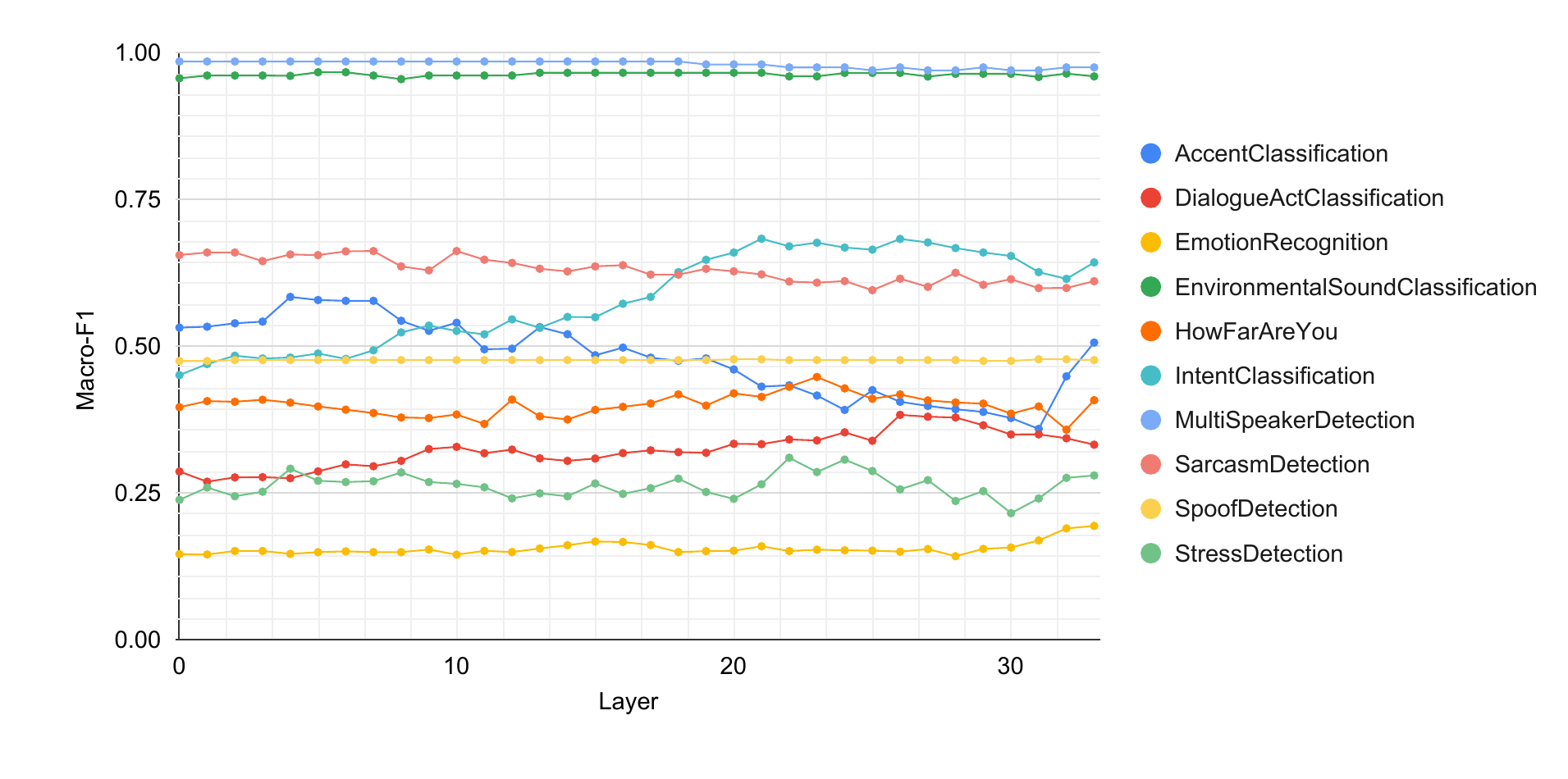}  
        \caption{Qwen2-Audio-7B}
        \label{fig:fifth}
    \end{subfigure}
    \hspace{0.03\textwidth}  
    \begin{subfigure}[b]{0.3\textwidth}
        \centering
        \includegraphics[width=\textwidth]{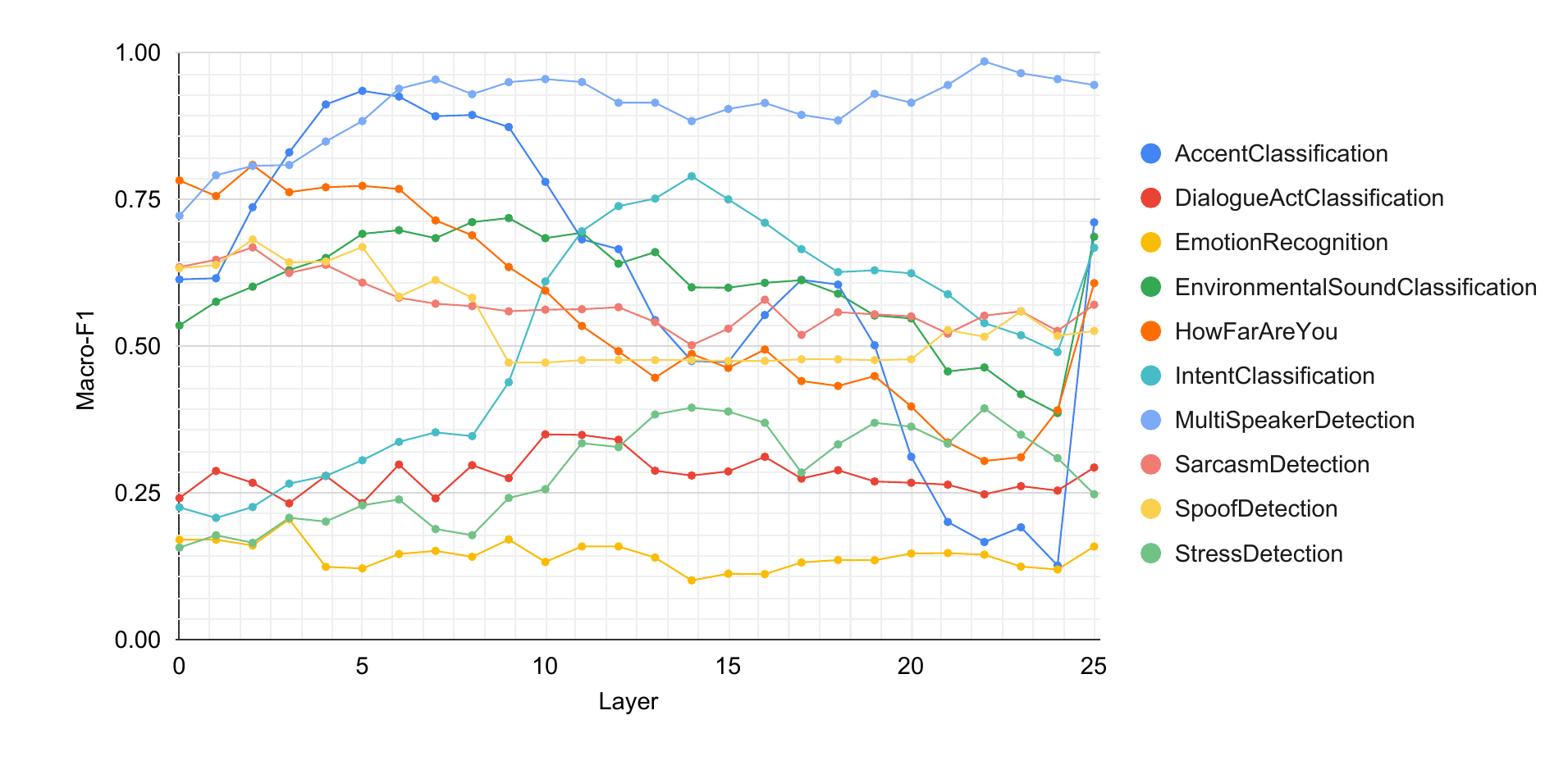}  
        \caption{Wav2Vec}
        \label{fig:sixth}
    \end{subfigure}

    \caption{Macro-F1 scores for KNN classifier for different models reported for different layers. We take encoder part of all models except Qwen which is decoder only model.}
    \label{fig:knn-macro-f1-across-layers}
\end{figure*}

From our experiments, we observe that \textit{Whisper} models show a convergence of performance, with lower-performing tasks like \textit{Stress Detection} and \textit{Intent Classification} gradually improving in deeper layers, while high-performing tasks like \textit{MultiSpeaker Detection} and \textit{Environmental Sound Classification} maintain strong performance. This suggests deeper layers in \textit{Whisper} learn generalized representations that improve complex tasks without negatively affecting simpler ones.

HuBERT and Wav2Vec perform comparably to Whisper, particularly excelling in \textit{MultiSpeaker Detection}. \textit{Wav2Vec} shows more layer-wise variability, potentially offering opportunities for task-specific layer selection. Both models improve significantly in Intent Classification in deeper layers, highlighting their strength in semantic tasks.

Qwen2-Audio models exhibit stability across layers, performing consistently well, especially in \textit{Environmental Sound Classification}. The Qwen2-Audio-Instruct variant shows improvement in \textit{Intent Classification} early on, likely due to instruction tuning.

Seamless-medium shows a linear F1 score increase across layers, while Seamless-v2-large remains stable but worse. \textit{Accent Classification} trends downwards in both models, indicating a possible trade-off between accent-specific features and general speech representations. 

\begin{figure}[!hbt]
    \centering
    \includegraphics[width=1.\linewidth]{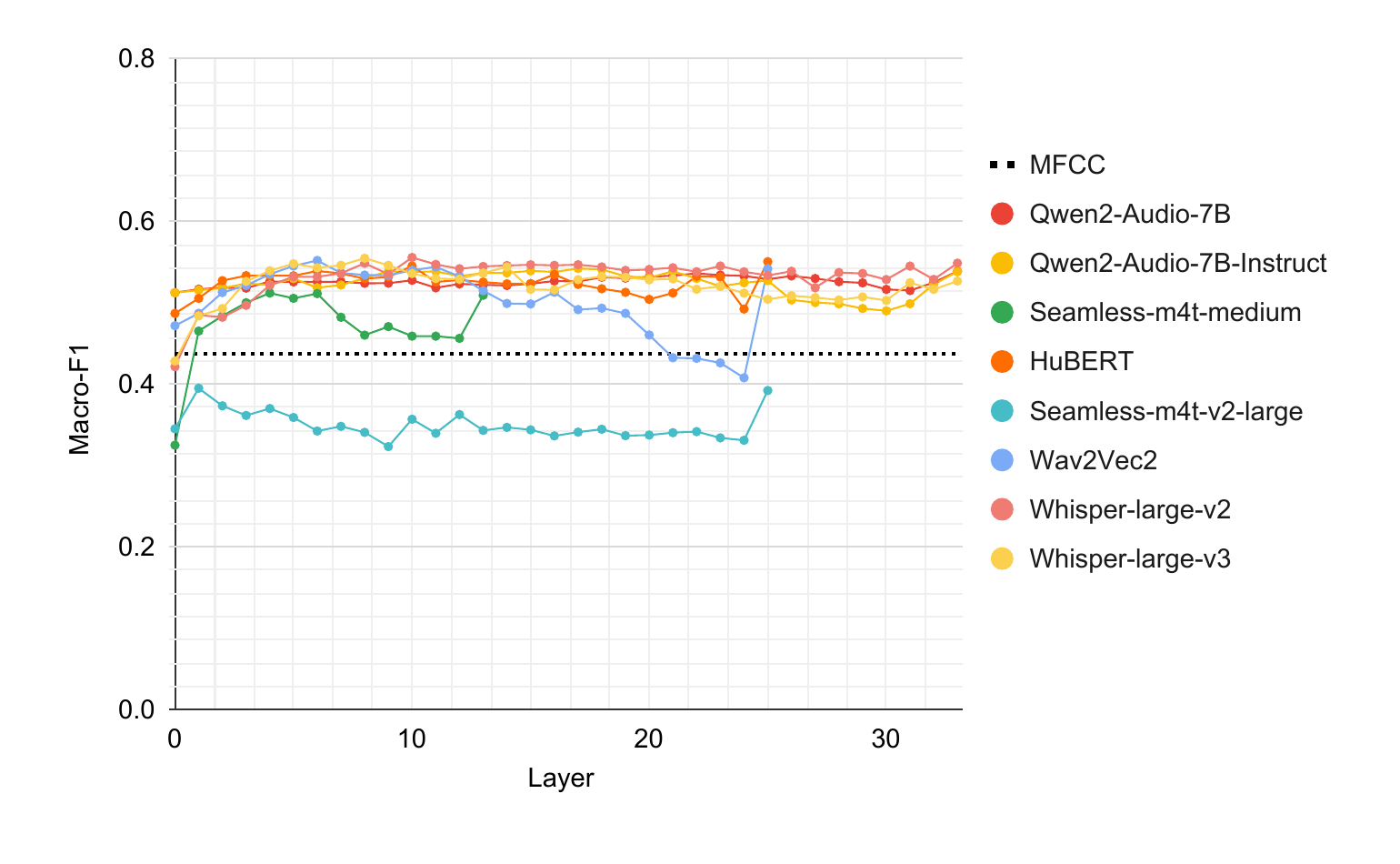}
   \caption{Macro-F1 scores for the KNN classifier, averaged across ten tasks for each layer of all models. The dotted line indicates the Macro-F1 score for MFCC features as a baseline.}
    \label{fig:task-average}
\end{figure}
We also experiment with handcrafted audio features from Librosa to provide a lower bound for the representations learned by the model. Our results show that nearly all models capture better audio features than the best Librosa feature (MFCC), as shown in Figure~\ref{fig:task-average}. Detailed results for the Librosa features are provided in Appendix~\ref{appendix-subsec:librosa-results}.

   

\subsection{Classifier Selection}
The distinction between KNN and models with learnable parameters (e.g., NN) highlights key differences in data interpretation. KNN represents classes as concatenations of convex polytopes, limiting its adaptability to complex decision boundaries, while NNs model arbitrary boundaries, allowing for more flexible and nuanced data interpretations. This flexibility explains NN's better performance across tasks, as deeper layers refine representations by aggregating categories into connected convex clusters, enhancing classification accuracy, particularly in tasks with subtle class distinctions like \textit{MultiSpeaker Detection} and \textit{Emotion Recognition}.

\begin{figure*}[!htb]
    \centering
    \begin{subfigure}[t]{1\textwidth}
        \centering
        \includegraphics[width=\linewidth]{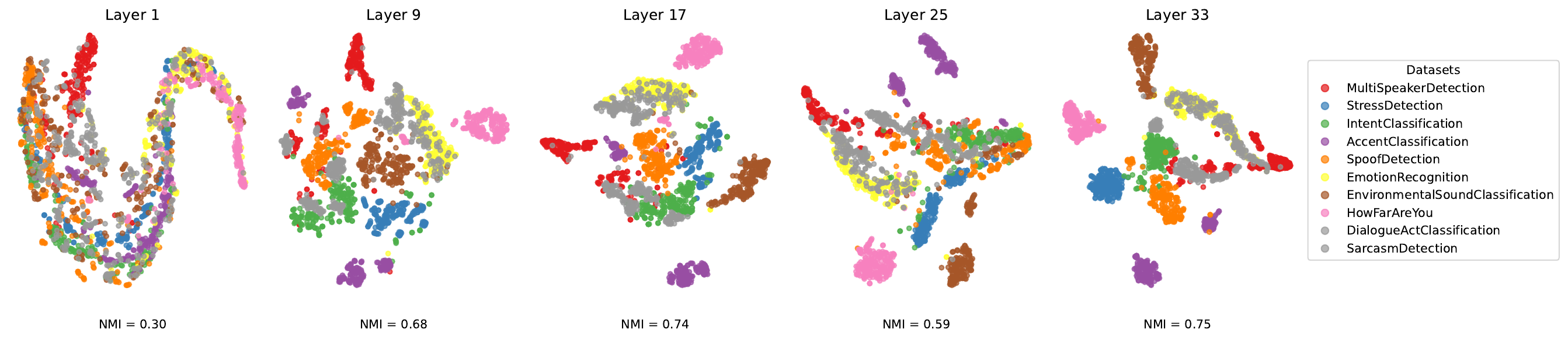}
        \caption{Whisper-large-v3}
        \label{fig:wav2vec2}
    \end{subfigure}
    
    \vspace{0.5em}  

    \begin{subfigure}[t]{1\textwidth}
        \centering
        \includegraphics[width=\linewidth]{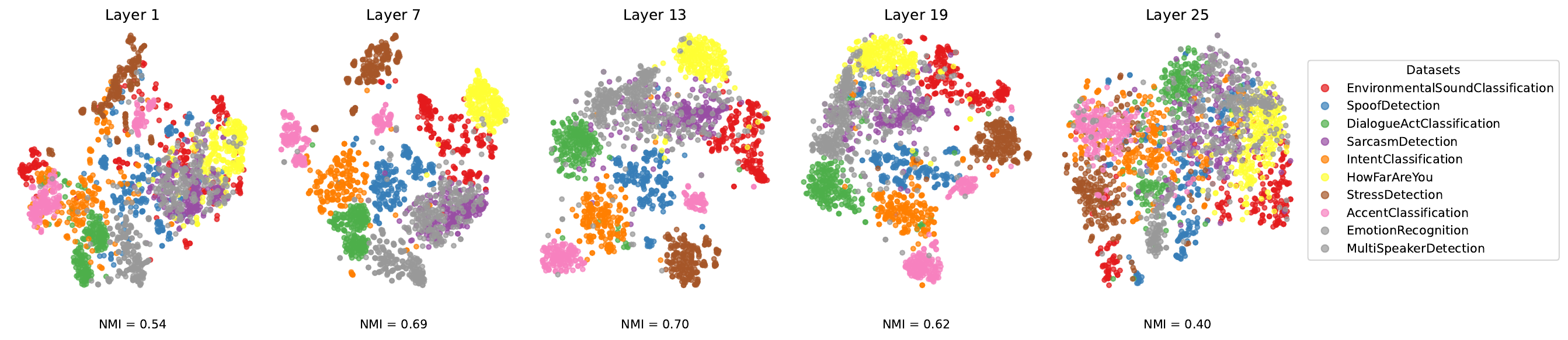}
        \caption{Wav2Vec}
        \label{fig:whisper}
    \end{subfigure}
    

    
    \caption{t-SNE embeddings for Wav2Vec and Whisper models.}
    \label{fig:tsne-plots}
\end{figure*}

Comparative analysis shows a clear advantage of Neural Networks (NNs) over KNNs when using features from HuBERT and Wav2Vec, especially in tasks requiring deeper audio understanding, such as \textit{Multi-speaker Detection} and \textit{Intent Classification}. NNs’ hierarchical nature enables them to exploit complex features, leading to superior performance.
Following~\cite{Jawahar2019WhatDB}, we study the features learned by speech foundation models. We quantify the mutual information using Normalized Mutual Information (NMI) and apply t-SNE to visualize the high-dimensional representations. We find that NMI increases in the intermediate layers of most models, indicating that these layers capture more task-relevant features as show in~\ref{fig:tsne-plots}.

\subsection{Layer-wise Performance Analysis}
For most models, performance improves in the initial layers before plateauing or slightly declining in later layers, indicating that earlier layers capture more generalizable features. Models like Whisper-large-v3 maintains consistent performance across layers, suggesting that features learned by these layers are robust. However, performance varies significantly across tasks—Accent Classification and MultiSpeaker Detection perform well, while Emotion and Stress Detection are more challenging for most models.

Models like HuBERT and Seamless-Medium show more fluctuations across layers, indicating uneven distribution of task-relevant features. Qwen2-Audio-Instruct exhibits stable performance across layers, hinting at a more distributed feature representation.

These findings underscore the variability in feature extraction across layers and tasks. Simpler tasks may benefit from earlier layers, while complex tasks often require features from later layers.

\subsection{The Role of Model Architecture and Pretraining in Representation Learning}
The observed patterns across model families emphasize the importance of architecture and pre-training in capturing task-relevant features. Whisper models, with their encoder-decoder architecture and large-scale supervised pre-training, show performance convergence across layers, gradually improving weaker tasks while maintaining strong ones.

In contrast, self-supervised models like \textit{HuBERT} and \textit{Wav2Vec} excel in speaker-related tasks due to their pre-training on unlabeled speech, effectively capturing fundamental speech characteristics. \textit{Qwen2-Audio} models exhibit stability across layers, excelling in intent classification, likely due to their multilingual, multitask pre-training. The performance difference between \textit{Seamless Medium} and \textit{Large} models highlights that scaling model size doesn't always result in uniform improvements across tasks and layers.

\section{Conclusion}\label{sec:conclusion}

We analyzed the feature representations learned by speech foundation models across various layers using ten tasks from the Dynamic-SUPERB benchmark. While models performed well on tasks like multi-speaker detection and accent classification, emotion recognition and stress detection remained challenging, with low F1 scores indicating difficulty in capturing emotional and contextual subtleties.

Our layer-wise analysis showed that earlier layers capture generalizable features, while mid and later layers contain more task-specific information. Models like Whisper and Qwen2-Audio demonstrated stable performance across layers, while HuBERT and Wav2Vec showed more variation. Strong zero-shot performance also correlated with better representation learning, highlighting the importance of architecture and pre-training.



\section{Limitations and Future Works}\label{sec:limitations}

While our study on speech foundation models provides valuable insights, it has several limitations. First, the Dynamic-SUPERB tasks selected may not comprehensively represent all speech-related challenges, particularly in diverse linguistic and contextual settings. Additionally, our zero-shot evaluations occasionally performed worse than random, suggesting the need for further investigation into prompt design and task framing. The choice of classifiers (K-Nearest Neighbors and Neural Networks) and their hyperparameters may have influenced the results, potentially not fully capturing the models' discriminative capabilities. Moreover, the real-world applicability of our findings may be constrained by factors such as data quality and environmental conditions. Future research should address these issues to gain a more complete understanding of model performance across broader tasks and environments.

\bibliography{custom}

\appendix

\section{Appendix}\label{sec:appendix}

\section{Experiments}~\label{appendix-sec:experiment}
\subsection{Tasks}~\label{appendix-subsec:tasks}

We choose ten tasks, broadly categorizing them into paralinguistic and non-paralinguistic groups. Each task in Dynamic-SUPERB comes with clearly defined instructions, facilitating the effective evaluation of model performance in previously unseen contexts.

\noindent\textbf{Dialogue Act Classification.} This task involves identifying the primary purpose of an utterance within its conversational context. Using the DailyTalk Dataset, the goal is to classify utterances into categories such as question, inform, directive, or commissive.

\noindent\textbf{Accent Classification.} This task focuses on recognizing and classifying different speech accents, utilizing the AccentDB Extended Dataset. Participants aim to accurately identify accents such as American, Australian, Bangla, British, Indian, Malayalam, Odiya, Telugu, or Welsh based on audio samples. 

\noindent\textbf{Emotion Recognition.} The objective here is to classify the emotional category of an utterance, using the Multimodal EmotionLines Dataset. This task can be challenging, as it often requires analyzing not just the linguistic content but also paralinguistic features like pitch and rhythm. 

\noindent\textbf{Stress Detection.} This task involves analyzing stress placement in English words using the MIR-SD dataset. The goal is to identify stress patterns, with possible responses ranging from zero to five, which is important for developing nuanced speech models.

\noindent\textbf{Environmental Sound Classification.} Focusing on recognizing and categorizing environmental sounds, this task employs the ESC50 dataset. Sounds are classified into five categories:

Animals (e.g., dog, cat)
Urban Noises (e.g., siren, chainsaw)
Human and Non-Speech Sounds (e.g., coughing, laughing)
Domestic Sounds (e.g., door knocks, vacuum cleaner)
Natural Soundscapes (e.g., rain, thunder)

\noindent\textbf{HowFarAreYou.} This task assesses the speaker's distance from a sound source using the 3DSpeaker dataset. Responses indicate distance in meters (e.g., 0.4m, 2.0m), providing insights into audio spatial characteristics critical for auditory scene analysis

\noindent\textbf{Intent Classification.} This task focuses on identifying the actionable item behind spoken messages, using the FluentSpeechCommands Dataset. The goal is to categorize intents such as activate, deactivate, or change language. 

\noindent\textbf{Sarcasm Detection.} This task aims to identify sarcasm or irony in speech audio, utilizing the MUStARD dataset. The objective is to determine the presence of sarcasm, with answers being either true or false.

\subsection{Speech Foundation Models}~\label{append-subsec:models}
\noindent\textbf{Qwen2-Audio~\cite{chu2024qwen2, chu2023qwen}.} Qwen-audio addresses the challenge of co-training multiple tasks and datasets through a multitask training framework. The model conditions the decoder output as hierarchical labels, working across more than 30 tasks, eight languages, and various audio types. It mitigates differences in task goals, languages, annotation granularity, and text structure by sharing labels. Qwen-Audio features an audio encoder and the Qwen-7B large language model, a 32-layer Transformer decoder with a hidden size of 4096, totaling 7.7 billion parameters. It employs a unified decoder framework, leveraging self-supervised learning on unlabeled data with task-specific adapters for downstream applications, such as ASR, TTS, voice conversion, and speech translation.

\noindent\textbf{HuBERT~\cite{hsu2021hubertselfsupervisedspeechrepresentation}.} HuBERT is pretrained using 960 hours of data from the LibriSpeech dataset. It is built on the BERT architecture with a multi-layer convolutional feature encoder and a Transformer network for contextualized representations. HuBERT learns discrete speech units iteratively by masking input features to predict hidden units and applying contrastive loss to enhance speech representation learning. The model is primarily used for ASR, speaker identification, emotion recognition, and speech enhancement.

\noindent\textbf{Wav2Vec~\cite{baevski2020wav2vec}.} Wav2Vec is pretrained on 960 hours of LibriSpeech data (base model) and 60,000 hours of Libri-Light data (large model). The model uses a convolutional feature encoder and a Transformer for context representation. It also includes a quantization module to learn discrete units. By masking latent representations and applying contrastive learning, Wav2Vec 2.0 identifies accurate speech representations. The model's downstream tasks include ASR, speaker identification, language identification, and emotion recognition.

\noindent\textbf{Seamless~\cite{barrault2023seamlessm4t}.} Seamless is trained on large-scale multilingual speech and text datasets, though specific details are not publicly disclosed. The model uses a modular approach with distinct components for recognition, translation, and synthesis, leveraging Transformer-based architectures. Seamless supports tasks such as speech-to-speech and speech-to-text translation, TTS, and ASR. It integrates these components for end-to-end optimization, sharing representations across modalities to enhance translation quality and naturalness.

\noindent\textbf{Whisper~\cite{radford2023robust}.} Whisper is trained on 680,000 hours of multilingual and multitask supervised data sourced from the web. The model employs an encoder-decoder Transformer architecture, with a convolutional layer for processing raw audio inputs. It uses multi-task training to handle a variety of tasks, including ASR, speech translation, language identification, and voice activity detection. Whisper processes raw audio with a convolutional network, encodes it into contextualized representations via a Transformer, and decodes it into textual outputs using task-specific tokens for different languages and tasks.

\section{Training}~\label{appendix-sec:training}
\subsection{Hyperparameters}~\label{appendix-subsec:hyperparameters}
\begin{table}[h!]
\centering
\begin{tabular}{lc}
\toprule
\textbf{Hyperparameter}    & \textbf{Value} \\  \midrule
Neighbors (\texttt{k}) & 5               \\ 
Folds & 5               \\ 
Metric          & Macro-F1        \\  \bottomrule
\end{tabular}
\caption{KNN Hyperparameters}

\label{tab:knn_hyperparams}
\end{table}

\begin{table}[h!]
\centering
\begin{tabular}{lc}
\toprule
\textbf{Hyperparameter}    & \textbf{Value}   \\ \midrule
Input Size           & $\in \mathbb{R}^n$ \\ 
Hidden Size          & 128             \\ 
Layers        & 2            \\ 
Activation        & ReLU            \\ 
Optimizer                  & Adam            \\ 
Learning Rate              & 0.001           \\ 
Loss              & Cross-Entropy \\ 
Batch Size                 & 16              \\ 
Epochs           & 20              \\ 
Folds & 5         \\ 
Metric          & Macro-F1        \\  \bottomrule
\end{tabular}
\caption{Hyperparameters for neural network classifier.}
\label{tab:dnn_hyperparams}
\end{table}

\section{Results}~\label{appendix-sec:results}
\subsection{Zero-shot Results}~\label{appendix-subsec:zero-shot-results}
\begin{table*}[htbp]
\centering
\small 
\setlength{\tabcolsep}{4pt} 
\begin{tabular}{cccccccccccc}
\toprule
\textbf{Model}                    & \textbf{Prompt}  & \rotatebox{0}{\textbf{Accent}} & \rotatebox{0}{\textbf{DialAct}} & \rotatebox{0}{\textbf{Emotion}} & \rotatebox{0}{\textbf{EnvSound}} & \rotatebox{0}{\textbf{HowFar}} & \rotatebox{0}{\textbf{Intent}} & \rotatebox{0}{\textbf{MSpeaker}} & \rotatebox{0}{\textbf{Sarcasm}} & \rotatebox{0}{\textbf{Spoof}} & \rotatebox{0}{\textbf{Stress}} \\ \midrule
Random  &   & 9.66  & 23.59 & 11.31  & 9.86  & 33.53   & 15.32 & 49.76 & 49.43   & 39.76 & 12.55 \\ 
\midrule
 & MCQ   & 5.54 & 27.89 & 6.70 & 6.97 & 34.50 & 27.51 & 37.11 & 43.98 & 52.99 & 9.02  \\ 
W-L-v2  & Quiz    & 9.96 & 25.68 & 7.35 & 10.21 & 25.39 & 25.83 & 33.33 & 47.87 & 52.00 & 11.32 \\ 
                  & Blank  & 9.60 & 28.10 & 7.62 & 10.69 & 23.76 & 24.68 & 34.20 & 44.61 & 50.07 & 11.78 \\ 
\hdashline
 & MCQ & 4.60 & 17.34 & 4.37 & 1.18 & 28.11 & 22.37 & 47.05 & 36.48 & 48.90 & 12.17 \\ 
W-L-v3 & Quiz   &  5.71 & 20.14 & 4.71 & 8.38 & 32.33 & 21.01 & 50.55 & 36.48 & 49.34 & 16.96  \\ 
                  & Blank & 5.59 & 17.34 & 4.85 & 4.10 & 27.78 & 16.45 & 50.60 & 36.03 & 52.28 & 14.31 \\ 
\hdashline
 & MCQ  & 8.81 & 21.74 & 5.33 & 14.92 & 25.99 & 29.29 & 58.00 & 51.46 & 54.40 & 10.76 \\ 
W-M.en & Quiz& 7.30 & 16.16 & 7.45 & 10.58 & 25.14 & 21.89 & 51.86 & 48.30 & 55.84 & 10.90 \\ 
                   & Blank & 7.94 & 16.01 & 4.35 & 19.67 & 28.33 & 21.15 & 48.26 & 47.59 & 54.21 & 9.26 \\ 
\hdashline
& MCQ  & 3.84 & 16.30 & 8.52 & 3.57 & 29.75 & 29.72 & 36.77 & 45.45 & 50.28 & 11.33 \\ 
W-M  & Quiz  & 6.79 & 15.12 & 9.97 & 4.98 & 33.23 & 24.65 & 37.36 & 45.36 & 46.67 & 14.63 \\ 
                   & Blank & 8.17 & 18.13 & 5.97 & 16.27 & 30.39 & 23.49 & 33.49 & 44.15 & 46.32 & 14.00 \\
\hdashline
& MCQ  & 10.57 & 23.27 & 7.33 & 9.93 & 28.41 & 35.61 & 45.91 & 42.86 & 48.62 & 13.84\\ 
SM4T-M  & Quiz  &10.57 & 23.27 & 7.33 & 9.93 & 28.41 & 35.61 & 45.91 & 42.86 & 48.62 & 13.84\\ 
                   & Blank & 10.57 & 23.27 & 7.33 & 9.93 & 28.41 & 35.61 & 45.91 & 42.86 & 48.62 & 13.84 \\
                   \hdashline

                   & MCQ  & 1.55 & 4.34 & 3.83 & 1.74 & 16.16 & 2.18 & 33.77 & 35.06 & 47.78 & 1.74  \\ 
SM4T-v2-L  & Quiz  & 1.55 & 7.26 & 2.83 & 1.74 & 16.54 & 7.49 & 32.89 & 35.06 & 47.78 & 1.74\\ 
                   & Blank & 0.75 & 15.87 & 1.10 & 1.48 & 17.28 & 2.18 & 33.77 & 35.06 & 7.83 & 4.22 \\
                   \hdashline
& MCQ   & 2.21 & 23.33 & 29.90 & 30.11 & 24.60 & 44.52 & 32.89 & 35.06 & 47.64 & 10.53   \\
        Qwen2  & Quiz  & 2.11 & 25.99 & 2.71 & 10.19 & 30.29 & 16.79 & 84.01 & 35.06 & 47.78 & 0.34  \\
               & Blank & 7.51 & 29.80 & 26.49 & 45.07 & 19.92 & 29.09 & 33.77 & 35.06 & 47.37 & 17.38 \\
        \hdashline
        & MCQ   & 4.05 & 33.15 & 12.83 & 3.38 & 20.23 & 27.20 & 32.89 & 37.71 & 7.83 & 0.35   \\
        Qwen2-I & Quiz  & 6.40 & 16.83 & 10.01 & 2.23 & 26.68 & 23.89 & 32.89 & 45.55 & 9.56 & 9.55  \\
               & Blank & 1.68 & 21.96 & 11.76 & 14.98 & 31.76 & 18.62 & 58.50 & 41.69 & 7.83 & 8.86   \\
\bottomrule
\end{tabular}
\caption{Zero-shot results for different prompts. Abbreviations: W - Whisper, L - Large, D - DistilWhisper, M - Medium, SM4T - SeamlessM4T, I - Instruct.}
\label{tab:speech-model-performance}
\end{table*}

\subsection{Librosa Features Results}~\label{appendix-subsec:librosa-results}
\begin{table*}[ht]
\centering
\resizebox{\textwidth}{!}{%
\begin{tabular}{lcccccccccc}
\toprule
\textbf{Feature} & \textbf{Accent} & \textbf{DialAct} & \textbf{Emotion} & \textbf{EnvSound} & \textbf{Distance} & \textbf{Intent} & \textbf{MSpeaker} & \textbf{Sarcasm} & \textbf{Spoof} & \textbf{Stress} \\
\midrule
MFCC & 27.8$^{6.1}$ / 44.2$^{6.9}$ & 54.9$^{9.1}$ / 68.0$^{13.7}$ & 60.0$^{16.3}$ / 75.3$^{10.6}$ & 16.5$^{5.6}$ / 23.3$^{4.6}$ & 73.3$^{3.6}$ / 90.0$^{2.5}$ & 83.3$^{11.1}$ / 85.3$^{7.3}$ & 28.0$^{6.6}$ / 29.6$^{5.3}$ & 59.5$^{11.8}$ / 71.6$^{3.0}$ & 15.7$^{2.2}$ / 20.8$^{0.9}$ & 17.6$^{4.2}$ / 17.5$^{3.2}$ \\
Mel-Spectrogram & 27.8$^{7.4}$ / 40.6$^{6.9}$ & 47.5$^{0.5}$ / 47.8$^{0.3}$ & 52.9$^{4.6}$ / 63.3$^{2.1}$ & 18.0$^{5.2}$ / 27.7$^{2.5}$ & 62.4$^{7.6}$ / 57.2$^{5.5}$ & 47.0$^{6.0}$ / 65.5$^{8.3}$ & 24.1$^{6.8}$ / 27.3$^{5.3}$ & 52.1$^{6.3}$ / 61.8$^{5.1}$ & 22.0$^{6.9}$ / 23.2$^{6.1}$ & 12.6$^{3.8}$ / 14.5$^{3.0}$ \\
Chroma CENS & 23.2$^{4.0}$ / 30.4$^{5.7}$ & 51.6$^{8.2}$ / 47.8$^{0.3}$ & 54.4$^{6.7}$ / 59.8$^{5.2}$ & 11.7$^{4.7}$ / 19.0$^{4.5}$ & 49.8$^{4.8}$ / 56.4$^{2.5}$ & 43.3$^{7.6}$ / 39.5$^{8.9}$ & 21.5$^{7.3}$ / 24.8$^{4.8}$ & 39.6$^{6.8}$ / 57.9$^{3.8}$ & 26.0$^{9.5}$ / 21.1$^{2.8}$ & 15.4$^{3.7}$ / 10.7$^{1.1}$ \\
\bottomrule
\end{tabular}%
}
\caption{Performance of KNN / DNN classifiers across tasks using MFCC, Mel Spectrogram, and Chroma CENS features. We report five folds cross-validation scores, with standard deviations shown as superscripts.}
\end{table*}

\subsection{Classificatin Results on Layer-wise Features}\label{appendix-subsec:layerwise}
\begin{table*}[ht]
\centering
\resizebox{\textwidth}{!}{%
\begin{tabular}{lcccccccccc}
\toprule
\textbf{Layer} & \textbf{Accent} & \textbf{DialAct} & \textbf{Emotion} & \textbf{EnvSound} & \textbf{Distance} & \textbf{Intent} & \textbf{MSpeaker} & \textbf{Sarcasm} & \textbf{Spoof} & \textbf{Stress} \\
\midrule
En-0 & 43.5$^{5.4}$ / 86.3$^{3.7}$ & 25.9$^{8.4}$ / 37.9$^{4.7}$ & 14.3$^{4.6}$ / 18.8$^{1.3}$ & 39.1$^{1.8}$ / 69.2$^{3.2}$ & 70.2$^{6.1}$ / 90.1$^{5.7}$ & 21.0$^{6.8}$ / 32.5$^{7.1}$ & 76.2$^{9.0}$ / 85.8$^{7.8}$ & 62.8$^{13.8}$ / 74.4$^{9.3}$ & 47.6$^{0.3}$ / 79.5$^{6.9}$ & 20.5$^{5.0}$ / 25.8$^{5.5}$ \\
En-1 & 63.1$^{7.5}$ / 84.8$^{6.4}$ & 34.5$^{10.2}$ / 36.4$^{4.2}$ & 18.1$^{4.1}$ / 25.7$^{6.7}$ & 60.4$^{3.6}$ / 81.5$^{3.4}$ & 69.8$^{4.5}$ / 88.6$^{3.9}$ & 20.0$^{4.6}$ / 31.9$^{4.0}$ & 76.4$^{9.2}$ / 86.4$^{8.8}$ & 64.1$^{11.7}$ / 76.1$^{11.1}$ & 57.9$^{12.9}$ / 84.3$^{9.5}$ & 20.3$^{5.0}$ / 26.9$^{6.3}$ \\
En-2 & 62.7$^{6.7}$ / 83.3$^{6.3}$ & 32.3$^{9.8}$ / 38.4$^{2.2}$ & 16.7$^{5.6}$ / 20.8$^{3.7}$ & 60.6$^{5.6}$ / 81.6$^{4.3}$ & 71.3$^{4.9}$ / 87.7$^{4.7}$ & 20.1$^{5.7}$ / 32.5$^{4.7}$ & 77.6$^{8.2}$ / 91.8$^{6.2}$ & 60.5$^{8.4}$ / 74.3$^{11.6}$ & 56.8$^{11.7}$ / 76.5$^{8.6}$ & 23.5$^{4.6}$ / 26.7$^{6.9}$ \\
En-3 & 73.2$^{6.9}$ / 85.3$^{5.5}$ & 24.4$^{6.4}$ / 39.8$^{3.9}$ & 13.4$^{1.5}$ / 20.3$^{4.3}$ & 68.6$^{7.9}$ / 85.5$^{3.2}$ & 76.9$^{4.4}$ / 89.6$^{1.0}$ & 23.9$^{4.2}$ / 35.5$^{4.7}$ & 82.3$^{9.3}$ / 92.3$^{4.7}$ & 58.7$^{13.2}$ / 71.7$^{9.5}$ & 57.6$^{13.8}$ / 79.0$^{6.0}$ & 17.8$^{3.2}$ / 30.1$^{3.6}$ \\
En-4 & 75.9$^{4.3}$ / 91.2$^{3.8}$ & 31.1$^{7.3}$ / 37.5$^{3.7}$ & 12.5$^{1.9}$ / 25.2$^{10.3}$ & 70.7$^{6.3}$ / 87.7$^{2.9}$ & 82.4$^{5.6}$ / 88.2$^{2.5}$ & 23.9$^{4.8}$ / 37.2$^{1.1}$ & 82.9$^{9.0}$ / 95.4$^{4.7}$ & 58.9$^{11.8}$ / 71.6$^{12.2}$ & 66.2$^{12.0}$ / 70.7$^{12.5}$ & 16.8$^{2.4}$ / 29.7$^{3.2}$ \\
\bottomrule
\end{tabular}
}
\caption{Whisper-Large-v2}
\end{table*}

\begin{table*}[ht]
\centering
\resizebox{\textwidth}{!}{%
\begin{tabular}{lcccccccccc}
\toprule
\textbf{Layer} & \textbf{Accent} & \textbf{DialAct} & \textbf{Emotion} & \textbf{EnvSound} & \textbf{Distance} & \textbf{Intent} & \textbf{MSpeaker} & \textbf{Sarcasm} & \textbf{Spoof} & \textbf{Stress} \\
\midrule
En-0 & 51.2$^{7.8}$ / 84.6$^{6.8}$ & 31.3$^{9.2}$ / 35.9$^{4.9}$ & 15.1$^{2.7}$ / 20.7$^{2.6}$ & 43.8$^{5.9}$ / 66.4$^{2.8}$ & 62.8$^{9.4}$ / 88.0$^{2.8}$ & 17.6$^{2.5}$ / 31.1$^{6.0}$ & 71.1$^{10.6}$ / 87.4$^{7.1}$ & 60.1$^{6.7}$ / 72.8$^{9.7}$ & 52.8$^{10.5}$ / 76.7$^{4.8}$ & 22.4$^{7.3}$ / 24.4$^{5.3}$ \\
En-1 & 66.2$^{10.3}$ / 89.0$^{4.7}$ & 23.9$^{6.5}$ / 39.1$^{6.1}$ & 16.5$^{2.4}$ / 21.3$^{4.9}$ & 64.3$^{6.1}$ / 81.8$^{3.2}$ & 74.9$^{3.9}$ / 90.1$^{5.5}$ & 16.6$^{4.3}$ / 31.8$^{5.5}$ & 77.0$^{8.2}$ / 91.9$^{5.5}$ & 63.4$^{12.6}$ / 69.6$^{15.6}$ & 62.7$^{12.6}$ / 76.7$^{4.3}$ & 18.4$^{3.6}$ / 30.0$^{8.5}$ \\
En-2 & 78.4$^{6.0}$ / 92.2$^{5.0}$ & 25.8$^{5.2}$ / 36.3$^{9.0}$ & 13.5$^{1.1}$ / 18.9$^{2.9}$ & 68.9$^{3.6}$ / 84.2$^{2.4}$ & 81.7$^{4.8}$ / 93.6$^{3.3}$ & 20.0$^{5.8}$ / 38.5$^{4.2}$ & 77.2$^{8.8}$ / 93.9$^{5.4}$ & 61.0$^{14.8}$ / 72.6$^{13.4}$ & 47.2$^{0.5}$ / 83.3$^{5.8}$ & 19.1$^{4.2}$ / 30.8$^{7.9}$ \\
En-3 & 77.2$^{4.1}$ / 93.4$^{2.6}$ & 27.0$^{5.6}$ / 33.4$^{7.1}$ & 14.8$^{3.1}$ / 18.7$^{3.0}$ & 74.3$^{5.2}$ / 85.7$^{3.7}$ & 87.7$^{2.6}$ / 95.0$^{2.7}$ & 20.4$^{2.3}$ / 37.5$^{4.4}$ & 81.6$^{6.5}$ / 94.9$^{5.5}$ & 62.5$^{14.1}$ / 73.0$^{12.9}$ & 59.9$^{9.9}$ / 80.7$^{5.4}$ & 20.8$^{6.0}$ / 31.9$^{8.1}$ \\
En-4 & 84.6$^{7.6}$ / 90.9$^{6.1}$ & 23.6$^{5.5}$ / 33.3$^{5.1}$ & 13.4$^{3.1}$ / 19.4$^{3.7}$ & 73.6$^{4.8}$ / 87.4$^{4.3}$ & 86.1$^{7.0}$ / 95.5$^{2.9}$ & 25.4$^{4.7}$ / 35.9$^{5.2}$ & 93.5$^{3.8}$ / 99.5$^{1.0}$ & 61.0$^{12.4}$ / 74.8$^{11.1}$ & 55.9$^{9.9}$ / 84.9$^{3.6}$ & 22.0$^{6.1}$ / 30.3$^{9.2}$ \\
\bottomrule
\end{tabular}
}
\caption{Whisper-Large-v3}
\end{table*}

\begin{table*}[ht]
\centering
\resizebox{\textwidth}{!}{%
\begin{tabular}{lcccccccccc}
\toprule
\textbf{Layer} & \textbf{Accent} & \textbf{DialAct} & \textbf{Emotion} & \textbf{EnvSound} & \textbf{Distance} & \textbf{Intent} & \textbf{MSpeaker} & \textbf{Sarcasm} & \textbf{Spoof} & \textbf{Stress} \\
\midrule
En-0 & 59.3$^{8.0}$ / 87.4$^{3.1}$ & 29.4$^{7.2}$ / 35.1$^{5.0}$ & 14.9$^{3.0}$ / 20.9$^{2.6}$ & 43.7$^{1.5}$ / 72.5$^{4.6}$ & 69.8$^{4.8}$ / 91.6$^{4.0}$ & 17.1$^{6.3}$ / 32.2$^{5.6}$ & 75.6$^{12.7}$ / 86.8$^{5.5}$ & 59.8$^{8.7}$ / 72.4$^{10.2}$ & 51.6$^{8.2}$ / 73.9$^{14.0}$ & 21.1$^{3.6}$ / 26.5$^{6.3}$ \\
En-1 & 68.4$^{10.5}$ / 86.6$^{8.0}$ & 28.8$^{4.4}$ / 40.4$^{6.3}$ & 14.4$^{9.3}$ / 18.8$^{6.5}$ & 59.1$^{7.9}$ / 79.4$^{4.0}$ & 79.5$^{2.8}$ / 89.0$^{1.2}$ & 21.8$^{3.2}$ / 30.7$^{4.5}$ & 77.4$^{6.2}$ / 93.9$^{3.9}$ & 60.6$^{11.3}$ / 67.2$^{12.1}$ & 47.6$^{0.3}$ / 81.3$^{9.3}$ & 24.4$^{3.6}$ / 28.2$^{5.7}$ \\
En-2 & 76.0$^{11.2}$ / 91.3$^{6.8}$ & 29.1$^{3.7}$ / 44.3$^{5.4}$ & 10.1$^{2.1}$ / 20.8$^{3.4}$ & 66.1$^{3.5}$ / 84.2$^{4.0}$ & 86.2$^{4.2}$ / 90.6$^{4.7}$ & 23.6$^{6.4}$ / 33.4$^{5.2}$ & 78.6$^{9.5}$ / 96.5$^{2.6}$ & 60.3$^{11.7}$ / 68.9$^{12.3}$ & 51.1$^{6.4}$ / 72.8$^{14.2}$ & 20.2$^{3.0}$ / 30.4$^{4.2}$ \\
En-3 & 87.4$^{7.5}$ / 91.3$^{6.6}$ & 27.5$^{3.9}$ / 42.9$^{3.2}$ & 11.8$^{6.6}$ / 19.1$^{2.1}$ & 77.5$^{4.2}$ / 86.5$^{5.5}$ & 87.0$^{4.1}$ / 94.5$^{2.5}$ & 24.1$^{8.2}$ / 36.1$^{6.5}$ & 82.7$^{7.6}$ / 95.8$^{7.2}$ & 57.8$^{15.4}$ / 67.6$^{10.9}$ & 51.9$^{8.1}$ / 81.8$^{5.0}$ & 20.4$^{4.0}$ / 28.3$^{6.3}$ \\
En-4 & 87.6$^{9.0}$ / 94.4$^{5.6}$ & 31.4$^{2.2}$ / 36.7$^{6.7}$ & 12.3$^{2.7}$ / 18.6$^{2.7}$ & 79.8$^{3.3}$ / 88.9$^{4.1}$ & 84.7$^{3.5}$ / 91.1$^{2.5}$ & 30.1$^{2.9}$ / 35.9$^{5.0}$ & 91.9$^{5.3}$ / 98.0$^{3.0}$ & 60.2$^{11.9}$ / 72.2$^{13.1}$ & 58.0$^{12.8}$ / 84.8$^{8.6}$ & 21.9$^{2.9}$ / 27.5$^{5.8}$ \\
\bottomrule
\end{tabular}
}
\caption{Whisper-Medium}
\end{table*}
\begin{table*}[ht]
\centering
\resizebox{\textwidth}{!}{%
\begin{tabular}{lcccccccccc}
\toprule
\textbf{Layer} & \textbf{Accent} & \textbf{DialAct} & \textbf{Emotion} & \textbf{EnvSound} & \textbf{Distance} & \textbf{Intent} & \textbf{MSpeaker} & \textbf{Sarcasm} & \textbf{Spoof} & \textbf{Stress} \\
\midrule
En-0 & 56.9$^{9.7}$ / 90.0$^{2.3}$ & 30.1$^{6.1}$ / 36.7$^{4.4}$ & 12.5$^{2.8}$ / 21.2$^{2.2}$ & 42.9$^{2.5}$ / 69.7$^{3.3}$ & 66.7$^{5.3}$ / 89.1$^{2.9}$ & 18.9$^{7.2}$ / 33.3$^{7.5}$ & 73.1$^{10.0}$ / 88.2$^{4.9}$ & 56.6$^{8.1}$ / 72.4$^{10.2}$ & 52.9$^{10.4}$ / 79.9$^{6.6}$ & 21.2$^{4.3}$ / 27.0$^{6.5}$ \\
En-1 & 66.7$^{9.9}$ / 86.1$^{12.0}$ & 32.3$^{9.0}$ / 38.1$^{8.5}$ & 12.5$^{0.9}$ / 22.4$^{5.3}$ & 58.6$^{5.0}$ / 75.1$^{4.3}$ & 79.4$^{1.8}$ / 89.1$^{3.8}$ & 14.3$^{3.3}$ / 31.4$^{3.3}$ & 80.2$^{7.1}$ / 90.8$^{4.2}$ & 55.7$^{12.0}$ / 69.8$^{13.6}$ & 58.0$^{12.8}$ / 80.3$^{17.4}$ & 19.5$^{6.1}$ / 29.8$^{8.0}$ \\
En-2 & 71.4$^{10.4}$ / 89.3$^{11.1}$ & 28.0$^{4.0}$ / 36.7$^{7.4}$ & 11.3$^{3.2}$ / 22.2$^{3.1}$ & 62.1$^{5.9}$ / 78.0$^{2.9}$ & 81.7$^{4.8}$ / 92.6$^{2.1}$ & 20.6$^{6.1}$ / 33.5$^{3.1}$ & 82.3$^{6.2}$ / 94.9$^{4.3}$ & 57.9$^{9.6}$ / 67.0$^{13.6}$ & 59.7$^{16.9}$ / 79.5$^{5.0}$ & 22.8$^{7.4}$ / 31.4$^{7.5}$ \\
En-3 & 78.4$^{12.7}$ / 92.0$^{7.6}$ & 29.2$^{5.7}$ / 39.6$^{7.2}$ & 10.6$^{1.3}$ / 18.6$^{3.2}$ & 63.9$^{4.8}$ / 81.0$^{1.5}$ & 86.5$^{5.0}$ / 95.6$^{3.6}$ & 16.0$^{2.3}$ / 32.1$^{3.9}$ & 80.6$^{8.8}$ / 95.3$^{7.0}$ & 63.2$^{13.8}$ / 70.5$^{13.3}$ & 51.0$^{6.5}$ / 77.5$^{15.0}$ & 21.7$^{6.0}$ / 30.4$^{6.9}$ \\
En-4 & 71.6$^{6.8}$ / 89.4$^{5.9}$ & 38.8$^{8.6}$ / 37.4$^{6.1}$ & 13.0$^{2.1}$ / 18.4$^{2.9}$ & 67.7$^{4.3}$ / 82.5$^{3.4}$ & 85.2$^{2.6}$ / 94.0$^{4.7}$ & 20.6$^{3.2}$ / 29.8$^{2.5}$ & 82.2$^{9.5}$ / 95.9$^{5.3}$ & 62.8$^{15.1}$ / 71.4$^{11.3}$ & 51.6$^{8.2}$ / 80.7$^{16.8}$ & 20.8$^{7.2}$ / 27.3$^{8.3}$ \\
\bottomrule
\end{tabular}
}
\caption{Whisper-Medium.en}
\end{table*}

\begin{table*}[ht]
\centering
\resizebox{\textwidth}{!}{%
\begin{tabular}{lcccccccccc}
\toprule
\textbf{Layer} & \textbf{Accent} & \textbf{DialAct} & \textbf{Emotion} & \textbf{EnvSound} & \textbf{Distance} & \textbf{Intent} & \textbf{MSpeaker} & \textbf{Sarcasm} & \textbf{Spoof} & \textbf{Stress} \\
\midrule
En-0 & 66.9$^{5.1}$ / 69.3$^{9.5}$ & 26.5$^{3.2}$ / 33.9$^{5.3}$ & 11.1$^{2.6}$ / 17.0$^{5.3}$ & 19.2$^{5.8}$ / 35.3$^{3.9}$ & 26.2$^{4.1}$ / 45.1$^{3.0}$ & 16.1$^{0.8}$ / 22.2$^{5.0}$ & 44.4$^{6.8}$ / 65.3$^{6.5}$ & 46.2$^{4.4}$ / 63.6$^{9.3}$ & 47.8$^{0.3}$ / 63.0$^{12.5}$ & 20.4$^{7.0}$ / 21.4$^{6.8}$ \\
En-1 & 75.5$^{2.6}$ / 81.8$^{3.2}$ & 24.0$^{3.4}$ / 34.9$^{5.9}$ & 14.1$^{2.3}$ / 20.4$^{6.1}$ & 54.5$^{5.0}$ / 79.3$^{4.9}$ & 66.4$^{7.9}$ / 81.7$^{3.9}$ & 26.3$^{5.2}$ / 35.2$^{11.4}$ & 77.1$^{7.8}$ / 95.4$^{3.8}$ & 57.1$^{7.8}$ / 70.9$^{8.5}$ & 47.6$^{0.3}$ / 84.6$^{6.1}$ & 22.5$^{4.2}$ / 28.5$^{3.8}$ \\
En-2 & 73.2$^{1.9}$ / 81.9$^{9.9}$ & 24.0$^{6.7}$ / 39.1$^{7.1}$ & 17.0$^{4.3}$ / 23.9$^{8.9}$ & 63.1$^{7.0}$ / 77.2$^{6.9}$ & 61.6$^{4.6}$ / 82.2$^{5.6}$ & 29.5$^{2.2}$ / 45.5$^{3.6}$ & 90.9$^{6.8}$ / 96.4$^{5.1}$ & 53.4$^{6.9}$ / 70.3$^{7.2}$ & 47.5$^{0.5}$ / 71.4$^{13.6}$ & 23.4$^{5.0}$ / 28.5$^{4.8}$ \\
En-3 & 84.4$^{2.6}$ / 86.9$^{8.4}$ & 23.9$^{5.5}$ / 35.9$^{8.9}$ & 18.1$^{4.9}$ / 19.7$^{4.2}$ & 58.8$^{5.5}$ / 75.9$^{7.3}$ & 61.8$^{5.6}$ / 86.0$^{5.3}$ & 28.8$^{4.4}$ / 47.0$^{5.2}$ & 93.9$^{5.3}$ / 95.9$^{4.2}$ & 60.1$^{4.2}$ / 78.1$^{2.9}$ & 47.2$^{0.5}$ / 74.5$^{14.0}$ & 22.7$^{6.4}$ / 30.7$^{2.6}$ \\
En-4 & 80.7$^{10.6}$ / 90.5$^{6.5}$ & 22.9$^{6.0}$ / 38.1$^{7.2}$ & 17.2$^{9.3}$ / 18.8$^{5.0}$ & 67.3$^{8.7}$ / 80.3$^{9.6}$ & 63.0$^{2.9}$ / 81.4$^{6.0}$ & 35.8$^{7.1}$ / 56.8$^{9.2}$ & 91.9$^{6.1}$ / 96.5$^{3.0}$ & 57.4$^{5.4}$ / 74.6$^{7.5}$ & 50.3$^{5.3}$ / 83.1$^{9.3}$ & 25.1$^{6.0}$ / 34.4$^{5.8}$ \\
\bottomrule
\end{tabular}
}
\caption{Seamless-Medium}
\end{table*}

\begin{table*}[ht]
\centering
\resizebox{\textwidth}{!}{%
\begin{tabular}{lcccccccccc}
\toprule
\textbf{Layer} & \textbf{Accent} & \textbf{DialAct} & \textbf{Emotion} & \textbf{EnvSound} & \textbf{Distance} & \textbf{Intent} & \textbf{MSpeaker} & \textbf{Sarcasm} & \textbf{Spoof} & \textbf{Stress} \\
\midrule
En-0 & 69.1$^{10.6}$ / 74.4$^{7.1}$ & 25.0$^{6.3}$ / 35.4$^{6.8}$ & 14.2$^{4.4}$ / 15.3$^{2.3}$ & 18.2$^{3.2}$ / 37.0$^{6.0}$ & 28.9$^{3.4}$ / 45.7$^{7.1}$ & 19.9$^{6.5}$ / 21.7$^{6.2}$ & 45.3$^{7.1}$ / 65.4$^{6.6}$ & 50.5$^{6.3}$ / 64.1$^{5.1}$ & 47.6$^{0.3}$ / 68.5$^{10.7}$ & 26.3$^{8.1}$ / 22.3$^{6.1}$ \\
En-1 & 56.0$^{5.7}$ / 72.9$^{4.8}$ & 22.7$^{5.3}$ / 34.7$^{4.9}$ & 10.6$^{1.8}$ / 12.9$^{2.7}$ & 51.9$^{3.1}$ / 71.3$^{5.7}$ & 21.3$^{5.6}$ / 30.7$^{3.2}$ & 19.4$^{2.1}$ / 26.9$^{7.2}$ & 73.4$^{9.9}$ / 91.9$^{7.5}$ & 53.9$^{3.3}$ / 68.2$^{7.4}$ & 60.9$^{11.2}$ / 76.0$^{17.5}$ & 24.8$^{6.2}$ / 25.1$^{3.4}$ \\
En-2 & 36.1$^{7.5}$ / 52.1$^{7.0}$ & 24.2$^{4.6}$ / 38.0$^{7.6}$ & 10.5$^{1.9}$ / 12.4$^{2.1}$ & 54.2$^{3.9}$ / 74.1$^{5.8}$ & 21.4$^{3.3}$ / 28.5$^{2.9}$ & 18.2$^{3.9}$ / 28.1$^{5.6}$ & 74.6$^{10.9}$ / 83.6$^{15.1}$ & 60.4$^{5.0}$ / 73.7$^{5.6}$ & 51.6$^{8.2}$ / 72.0$^{15.3}$ & 21.9$^{5.0}$ / 26.0$^{7.3}$ \\
En-3 & 29.5$^{3.5}$ / 45.6$^{11.6}$ & 21.7$^{4.7}$ / 36.2$^{2.3}$ & 10.5$^{1.9}$ / 14.7$^{2.7}$ & 57.0$^{5.6}$ / 75.6$^{6.0}$ & 22.7$^{5.0}$ / 30.3$^{2.7}$ & 23.1$^{3.9}$ / 29.8$^{3.4}$ & 70.6$^{11.4}$ / 80.5$^{14.1}$ & 58.7$^{3.7}$ / 71.1$^{6.8}$ & 47.4$^{0.6}$ / 74.6$^{10.2}$ & 20.3$^{5.3}$ / 26.4$^{3.0}$ \\
En-4 & 28.9$^{5.3}$ / 35.2$^{6.7}$ & 22.7$^{10.5}$ / 32.7$^{5.1}$ & 10.5$^{1.9}$ / 14.6$^{2.0}$ & 56.4$^{6.8}$ / 73.9$^{5.4}$ & 20.5$^{4.2}$ / 30.2$^{2.8}$ & 26.7$^{5.8}$ / 33.0$^{6.3}$ & 71.8$^{7.4}$ / 80.4$^{11.6}$ & 60.4$^{8.0}$ / 63.4$^{12.6}$ & 47.5$^{0.3}$ / 68.0$^{10.3}$ & 24.3$^{3.4}$ / 23.8$^{4.2}$ \\
\bottomrule
\end{tabular}
}
\caption{Seamless-Large}
\end{table*}

\begin{table*}[ht]
\centering
\resizebox{\textwidth}{!}{%
\begin{tabular}{lcccccccccc}
\toprule
\textbf{Layer} & \textbf{Accent} & \textbf{DialAct} & \textbf{Emotion} & \textbf{EnvSound} & \textbf{Distance} & \textbf{Intent} & \textbf{MSpeaker} & \textbf{Sarcasm} & \textbf{Spoof} & \textbf{Stress} \\
\midrule
En-0 & 61.4$^{3.9}$ / 87.7$^{5.5}$ & 24.1$^{9.5}$ / 37.9$^{4.1}$ & 17.1$^{4.1}$ / 22.8$^{3.7}$ & 53.5$^{10.3}$ / 71.1$^{7.2}$ & 78.2$^{6.8}$ / 85.6$^{3.0}$ & 22.6$^{8.6}$ / 33.3$^{5.0}$ & 72.2$^{8.3}$ / 91.7$^{7.3}$ & 63.5$^{5.2}$ / 72.7$^{6.2}$ & 63.3$^{12.8}$ / 77.6$^{15.8}$ & 15.7$^{5.4}$ / 19.6$^{3.7}$ \\
En-1 & 61.6$^{5.3}$ / 84.2$^{7.9}$ & 28.8$^{10.0}$ / 38.3$^{5.8}$ & 17.1$^{6.4}$ / 24.1$^{4.3}$ & 57.6$^{2.6}$ / 77.8$^{5.3}$ & 75.6$^{8.0}$ / 86.1$^{4.8}$ & 20.8$^{4.4}$ / 31.8$^{6.4}$ & 79.1$^{4.8}$ / 95.9$^{4.8}$ & 64.7$^{3.8}$ / 73.9$^{3.2}$ & 63.8$^{14.9}$ / 87.9$^{6.7}$ & 17.8$^{5.7}$ / 24.0$^{5.1}$ \\
En-2 & 73.7$^{3.9}$ / 94.5$^{4.4}$ & 26.8$^{7.8}$ / 35.5$^{4.6}$ & 16.1$^{5.0}$ / 19.8$^{5.2}$ & 60.1$^{6.8}$ / 81.2$^{6.0}$ & 80.9$^{4.2}$ / 85.5$^{3.9}$ & 22.6$^{6.1}$ / 37.2$^{9.5}$ & 80.7$^{6.1}$ / 95.9$^{4.8}$ & 66.8$^{3.8}$ / 73.7$^{5.4}$ & 68.2$^{13.5}$ / 84.4$^{19.4}$ & 16.5$^{5.2}$ / 23.4$^{4.6}$ \\
En-3 & 83.0$^{5.3}$ / 90.9$^{5.1}$ & 23.3$^{2.0}$ / 33.4$^{3.7}$ & 20.5$^{2.3}$ / 22.7$^{4.9}$ & 63.0$^{2.5}$ / 82.3$^{6.2}$ & 76.2$^{3.3}$ / 83.5$^{6.7}$ & 26.6$^{4.9}$ / 40.6$^{8.6}$ & 80.9$^{9.8}$ / 96.4$^{5.1}$ & 62.5$^{5.1}$ / 74.2$^{5.5}$ & 64.3$^{13.9}$ / 85.0$^{12.1}$ & 20.8$^{5.1}$ / 23.5$^{2.9}$ \\
En-4 & 91.2$^{3.7}$ / 96.7$^{2.8}$ & 27.9$^{7.8}$ / 38.4$^{3.4}$ & 12.4$^{2.2}$ / 23.0$^{5.6}$ & 65.0$^{6.0}$ / 82.4$^{3.5}$ & 77.1$^{0.9}$ / 82.5$^{6.3}$ & 27.9$^{5.7}$ / 44.7$^{11.5}$ & 84.9$^{6.2}$ / 96.9$^{5.0}$ & 63.9$^{5.5}$ / 70.5$^{6.7}$ & 64.4$^{13.7}$ / 82.0$^{18.1}$ & 20.2$^{5.8}$ / 24.7$^{4.0}$ \\
\bottomrule
\end{tabular}
}
\caption{Wav2Vec}
\end{table*}

\begin{table*}[ht]
\centering
\resizebox{\textwidth}{!}{%
\begin{tabular}{lcccccccccc}
\toprule
\textbf{Layer} & \textbf{Accent} & \textbf{DialAct} & \textbf{Emotion} & \textbf{EnvSound} & \textbf{Distance} & \textbf{Intent} & \textbf{MSpeaker} & \textbf{Sarcasm} & \textbf{Spoof} & \textbf{Stress} \\
\midrule
En-0 & 89.1$^{10.0}$ / 92.5$^{8.0}$ & 22.2$^{1.1}$ / 37.9$^{4.7}$ & 17.3$^{5.7}$ / 18.6$^{4.2}$ & 48.3$^{4.6}$ / 67.8$^{6.9}$ & 80.3$^{1.5}$ / 89.5$^{3.7}$ & 22.5$^{5.7}$ / 31.4$^{2.7}$ & 65.0$^{12.0}$ / 91.3$^{5.5}$ & 65.4$^{12.0}$ / 73.1$^{5.6}$ & 61.6$^{11.8}$ / 88.0$^{9.2}$ & 15.0$^{3.5}$ / 17.3$^{4.2}$ \\
En-1 & 85.9$^{7.4}$ / 96.1$^{4.0}$ & 25.2$^{5.2}$ / 38.6$^{5.8}$ & 15.5$^{9.1}$ / 20.7$^{3.0}$ & 56.3$^{4.4}$ / 74.3$^{4.3}$ & 75.0$^{3.5}$ / 89.0$^{4.5}$ & 20.0$^{5.8}$ / 35.7$^{6.0}$ & 79.6$^{4.4}$ / 95.4$^{5.8}$ & 61.2$^{8.4}$ / 77.8$^{4.8}$ & 70.0$^{14.1}$ / 88.4$^{3.4}$ & 16.7$^{5.1}$ / 24.3$^{2.4}$ \\
En-2 & 91.2$^{7.7}$ / 96.6$^{3.3}$ & 28.1$^{7.6}$ / 43.0$^{5.0}$ & 17.3$^{6.5}$ / 22.2$^{4.7}$ & 67.4$^{6.2}$ / 83.2$^{5.3}$ & 79.4$^{4.0}$ / 89.5$^{5.9}$ & 21.9$^{6.9}$ / 38.1$^{4.5}$ & 77.0$^{6.5}$ / 95.4$^{5.0}$ & 61.4$^{14.1}$ / 76.8$^{7.7}$ & 65.0$^{9.1}$ / 88.5$^{6.6}$ & 18.2$^{1.9}$ / 25.0$^{4.4}$ \\
En-3 & 90.1$^{7.5}$ / 96.1$^{4.3}$ & 25.8$^{8.1}$ / 35.4$^{1.9}$ & 17.1$^{3.7}$ / 21.2$^{3.9}$ & 71.6$^{6.9}$ / 85.5$^{6.7}$ & 79.3$^{3.3}$ / 87.6$^{5.7}$ & 29.0$^{5.6}$ / 38.6$^{6.2}$ & 78.0$^{7.9}$ / 97.0$^{3.8}$ & 61.3$^{13.7}$ / 73.2$^{7.8}$ & 61.0$^{11.1}$ / 83.3$^{5.8}$ & 19.9$^{4.1}$ / 25.4$^{4.2}$ \\
En-4 & 88.8$^{5.4}$ / 95.8$^{4.2}$ & 20.7$^{7.4}$ / 38.3$^{5.5}$ & 15.9$^{5.0}$ / 19.8$^{4.8}$ & 73.8$^{3.8}$ / 85.8$^{3.8}$ & 74.1$^{6.2}$ / 89.4$^{7.2}$ & 31.2$^{5.5}$ / 42.1$^{5.4}$ & 81.1$^{5.3}$ / 97.5$^{4.0}$ & 66.5$^{9.1}$ / 75.7$^{6.5}$ & 56.9$^{11.5}$ / 85.3$^{6.7}$ & 24.1$^{4.8}$ / 28.8$^{5.0}$ \\
\bottomrule
\end{tabular}
}
\caption{HuBERT}
\end{table*}

\begin{table*}[ht]
\centering
\resizebox{\textwidth}{!}{%
\begin{tabular}{lcccccccccc}
\toprule
\textbf{Layer} & \textbf{Accent} & \textbf{DialAct} & \textbf{Emotion} & \textbf{EnvSound} & \textbf{Distance} & \textbf{Intent} & \textbf{MSpeaker} & \textbf{Sarcasm} & \textbf{Spoof} & \textbf{Stress} \\
\midrule
De-0 & 53.2$^{4.7}$ / 78.3$^{6.5}$ & 28.7$^{4.2}$ / 51.5$^{6.1}$ & 14.6$^{2.4}$ / 23.5$^{4.6}$ & 95.6$^{3.2}$ / 96.6$^{2.7}$ & 39.6$^{5.1}$ / 58.7$^{2.7}$ & 45.1$^{8.9}$ / 74.3$^{3.4}$ & 98.5$^{1.2}$ / 98.5$^{2.0}$ & 65.5$^{9.5}$ / 76.1$^{10.1}$ & 47.5$^{0.8}$ / 59.1$^{9.4}$ & 23.8$^{3.9}$ / 35.9$^{6.6}$ \\
De-1 & 53.3$^{4.7}$ / 74.6$^{11.6}$ & 26.9$^{4.5}$ / 47.9$^{6.2}$ & 14.5$^{2.4}$ / 23.8$^{2.1}$ & 96.1$^{3.3}$ / 95.6$^{3.7}$ & 40.7$^{5.0}$ / 59.0$^{5.8}$ & 47.0$^{10.3}$ / 75.9$^{5.1}$ & 98.5$^{1.2}$ / 99.0$^{1.2}$ & 66.0$^{9.0}$ / 75.2$^{10.2}$ & 47.5$^{0.8}$ / 63.6$^{8.8}$ & 25.9$^{3.2}$ / 37.0$^{5.1}$ \\
De-2 & 53.9$^{4.5}$ / 78.3$^{8.1}$ & 27.7$^{5.1}$ / 50.5$^{5.1}$ & 15.1$^{3.0}$ / 24.0$^{3.6}$ & 96.1$^{3.3}$ / 98.1$^{1.7}$ & 40.6$^{6.8}$ / 60.2$^{3.4}$ & 48.4$^{8.8}$ / 74.1$^{6.8}$ & 98.5$^{1.2}$ / 99.5$^{1.0}$ & 66.0$^{9.0}$ / 75.1$^{8.9}$ & 47.6$^{0.5}$ / 55.6$^{10.2}$ & 24.5$^{3.5}$ / 38.1$^{5.3}$ \\
De-3 & 54.2$^{1.8}$ / 78.6$^{10.0}$ & 27.7$^{4.8}$ / 50.4$^{4.4}$ & 15.1$^{2.0}$ / 23.3$^{4.0}$ & 96.1$^{3.3}$ / 97.5$^{1.8}$ & 40.9$^{6.5}$ / 60.1$^{6.3}$ & 47.9$^{8.1}$ / 75.6$^{6.8}$ & 98.5$^{1.2}$ / 99.0$^{1.2}$ & 64.5$^{7.9}$ / 73.6$^{9.7}$ & 47.6$^{0.5}$ / 51.8$^{8.1}$ & 25.2$^{4.5}$ / 38.2$^{3.3}$ \\
De-4 & 58.4$^{4.1}$ / 76.3$^{9.3}$ & 27.5$^{3.5}$ / 51.1$^{6.0}$ & 14.6$^{1.7}$ / 22.2$^{4.6}$ & 96.0$^{2.5}$ / 97.5$^{2.3}$ & 40.4$^{6.6}$ / 57.1$^{4.7}$ & 48.1$^{9.8}$ / 78.6$^{5.6}$ & 98.5$^{1.2}$ / 99.5$^{1.0}$ & 65.6$^{7.4}$ / 76.2$^{10.8}$ & 47.6$^{0.5}$ / 55.5$^{10.2}$ & 29.1$^{8.0}$ / 37.8$^{5.9}$ \\
\bottomrule
\end{tabular}
}
\caption{Qwen2-Audio-7B}
\end{table*}

\begin{table*}[ht]
\centering
\resizebox{\textwidth}{!}{%
\begin{tabular}{lcccccccccc}
\toprule
\textbf{Layer} & \textbf{Accent} & \textbf{DialAct} & \textbf{Emotion} & \textbf{EnvSound} & \textbf{Distance} & \textbf{Intent} & \textbf{MSpeaker} & \textbf{Sarcasm} & \textbf{Spoof} & \textbf{Stress} \\
\midrule
De-0 & 53.2$^{4.7}$ / 78.6$^{9.9}$ & 28.7$^{4.2}$ / 50.7$^{3.8}$ & 14.6$^{2.4}$ / 24.5$^{5.5}$ & 95.6$^{3.2}$ / 97.0$^{2.5}$ & 39.6$^{5.1}$ / 57.4$^{3.2}$ & 45.1$^{8.9}$ / 78.5$^{5.3}$ & 98.5$^{1.2}$ / 99.5$^{1.0}$ & 65.5$^{9.5}$ / 73.3$^{10.9}$ & 47.5$^{0.8}$ / 51.6$^{8.6}$ & 23.8$^{3.9}$ / 36.1$^{3.5}$ \\
De-1 & 54.4$^{4.2}$ / 77.2$^{8.4}$ & 26.9$^{4.5}$ / 51.5$^{6.1}$ & 14.5$^{2.4}$ / 23.7$^{2.7}$ & 96.1$^{3.3}$ / 97.0$^{1.9}$ & 40.2$^{5.6}$ / 58.0$^{6.6}$ & 46.7$^{9.5}$ / 76.3$^{3.3}$ & 98.5$^{1.2}$ / 99.0$^{1.2}$ & 66.0$^{9.0}$ / 75.8$^{10.1}$ & 47.5$^{0.8}$ / 52.6$^{10.6}$ & 24.6$^{3.6}$ / 35.1$^{6.2}$ \\
De-2 & 53.3$^{4.7}$ / 75.3$^{8.6}$ & 28.3$^{4.5}$ / 49.2$^{7.9}$ & 15.1$^{3.0}$ / 23.9$^{3.7}$ & 96.1$^{3.3}$ / 97.9$^{2.1}$ & 40.6$^{6.8}$ / 57.6$^{4.1}$ & 48.2$^{8.6}$ / 76.0$^{4.1}$ & 98.5$^{1.2}$ / 99.0$^{1.2}$ & 65.4$^{8.6}$ / 75.0$^{9.0}$ & 47.6$^{0.5}$ / 51.0$^{6.5}$ & 24.5$^{3.5}$ / 38.4$^{5.0}$ \\
De-3 & 57.2$^{5.1}$ / 77.9$^{10.4}$ & 29.7$^{5.3}$ / 49.2$^{5.8}$ & 14.8$^{1.9}$ / 23.6$^{2.8}$ & 96.7$^{2.4}$ / 97.0$^{2.0}$ & 40.2$^{5.8}$ / 57.0$^{5.4}$ & 48.3$^{9.0}$ / 75.9$^{6.7}$ & 98.5$^{1.2}$ / 99.5$^{1.0}$ & 66.0$^{7.6}$ / 73.0$^{8.7}$ & 47.5$^{0.8}$ / 57.2$^{7.9}$ & 24.5$^{3.6}$ / 35.5$^{3.7}$ \\
De-4 & 56.6$^{5.6}$ / 77.3$^{8.8}$ & 29.3$^{4.9}$ / 49.5$^{5.5}$ & 15.4$^{2.4}$ / 25.0$^{2.7}$ & 96.0$^{2.5}$ / 97.6$^{2.5}$ & 39.1$^{6.1}$ / 56.2$^{5.8}$ & 48.0$^{10.0}$ / 77.3$^{7.6}$ & 98.5$^{1.2}$ / 99.0$^{1.2}$ & 66.0$^{8.4}$ / 72.7$^{11.1}$ & 47.6$^{0.5}$ / 58.8$^{9.8}$ & 25.2$^{3.0}$ / 36.3$^{5.0}$ \\
\bottomrule
\end{tabular}
}
\caption{Qwen2-Audio-7B-Instructs}
\end{table*}

\end{document}